\documentclass[preprint]{elsarticle}

\usepackage{amsmath,amsthm}
\usepackage{caption}
\usepackage{graphicx}
\usepackage{algpseudocode}
\usepackage{algorithm}
\usepackage{algpascal}
\usepackage{amssymb}
\usepackage{lineno,hyperref}
\modulolinenumbers[5]

\journal{Mechanism and Machine Theory}

\renewcommand{\figurename}{Fig.}
\begin{document}

\begin{frontmatter}



\author[Seyedamirtafrishi]{Seyed Amir Tafrishi}
\ead{amirtafrishi@yahoo.com}
\author[Seyedamirtafrishi]{Sandor M. Veres}
\ead{s.veres@sheffield.ac.uk}


\author[a2]{Esmaeil Esmaeilzadeh \corref{mycorrespondingauthor}}
\cortext[mycorrespondingauthor]{Corresponding author}
\ead{esmzadeh@tabrizu.ac.ir}

\author[a3]{Mikhail Svinin}
\ead{svinin@mech.kyushu-u.ac.jp}

\address[Seyedamirtafrishi]{Automation and Control Systems Department, University of Sheffield, Sheffield , UK.}
\address[a2]{Mechanical Engineering Department, University of Tabriz, Tabriz, Iran}
\address[a3]{Mechanical Engineering Department, Kyushu University, Fukuoka, Japan}

\title{Dynamical Behavior Investigation and Analysis of Novel Mechanism for Simulated Spherical Robot named "RollRoller"}




\begin{abstract}
This paper introduces a simulation study of fluid actuated multi-driven closed system as spherical mobile robot called "RollRoller". Robot's mechanism  design consists of two essential parts: tubes to lead a core and mechanical controlling parts to correspond movements. Our robot gets its motivation force by displacing the spherical movable mass known as core in curvy manners inside certain pipes. This simulation investigates by explaining the mechanical and structural features of the robot for creating hydraulic-base actuation via force and momentum analysis. Next, we categorize difficult and integrated 2D motions to omit unstable equilibrium points through derived nonlinear dynamics. We propose an algorithmic position control in forward direction that creates hybrid model as solution for motion planning problem in spherical robot. By deriving nonlinear dynamics of the spherical robot and implementing designed motion planning, we show how RollRoller can be efficient in high speed movements in comparison to the other pendulum-driven models. Then, we validate the results of this position control obtained by nonlinear dynamics via Adams/view simulation which uses the imported solid model of RollRoller. Lastly, We have a look to the circular maneuver of this robot by the same simulator.
\end{abstract}


\end{frontmatter}

\section{Introduction}
%
%
%
%

Mobile robots have involved in many new scientific discussions in planetary and field researches during these centuries. For instance, these robots can work in hard conditions at hazardous fields. Spherical mobile robots(SMR) as one of the most capable but highly complicated models have been being widely well-known among early researches and innovations for their specific characteristics. 
SMR's coverage as a ball-like symmetric shell gives diverse capabilities for overcoming limitations in robots with legs, wheels or other types, and makes it tolerable under serious environmentally problematic situation like humid, watery, or even toxic areas \cite{RollingNature}. As an initial motivation, NASA scholars developed SMR called Tumbleweeds rover for outer space explorations and they reason how it can be tolerable under challenging conditions and have economic advantages \cite{PolarNASA}.

\begin{figure*}
\centering
\includegraphics[width=4.8 in, height=1.2 in]{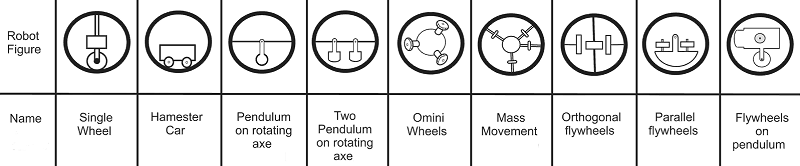}
\caption{Previous SMR models.}\label{Fig:PastSMRs}
\end{figure*}

The rolling sphere controllability was analyzed first theoretically in 1990 \cite{Tworigidrolling}. Haleme et al \cite{HalmeMotion1996} presented the first method for spherical robot motion control via kinematics, where a robot had mono-directional turning wheel (see Fig. \ref{Fig:PastSMRs}). Using the tiny car inside the ball shape shell was introduced, \cite{BicchiSpherenonholonomy} whereas, likelihood of losing control of the robot was considerably high. Also, this robot was consuming majority of space inside the ball.
After the torque-reaction base internal driving units, some models were developed to excite the SMRs by angular momentum forces. Robot containing rotating internal gyroscope with two motor drivers for each direction named Gyrover was invented \cite{BrownCMUGyrosignlewheel} as an example. Bhattacharya and Agrawal \cite{Bhattacharyarolling2000} also suggested perpendicular pair of rotors as alternative internal driving unit. The August robot was reproduced by Javadi in 2002 \cite{AugustJavadi2002}.  This model was making its motivation force solely through unbalancing the center of mass with replacing masses in different axis via rods. "Gyrosphere robot" \cite{MITSchroll2008} was introduced by Schroll from MIT in 2008 which it was utilizing both angular momentum and torque-reaction as hybrid system in SMR models. Although, pendulum-driven spherical robots have become common like BYQ-3 \cite{FirstPendulumBYQ2008} model in most known research areas, they could not manage to refine some crucial drawbacks such as limited velocity, unstable points and fully occupied inner space. VolVot \cite{VolvotSMR2011} and Omnicron \cite{Omnicron2012AIM} are the recently developed robots inspired by the previous SMR models to create hybrid locomotion forms. In addition, there have been many works in advancement of previous SMRs' motion planning by geometric phases \cite{GeometryPhase2005,GeometryPhase2014,GeometryPhase2015}, algorithm-base \cite{Algorithmbase2008,Algorithmbase2010} and other control methods \cite{Controlnonholonomic2012,Controlfuzzy2013}. The literature review of SMRs proves the discrete and perfunctory researches on various internal driving unit methodologies to actuate ball-shaped robots. This means there is still no unique designed model that satisfies all the characteristics for the ideal field research in current SMRs. 
\begin{figure}
\centering
\includegraphics[width=3.4 in]{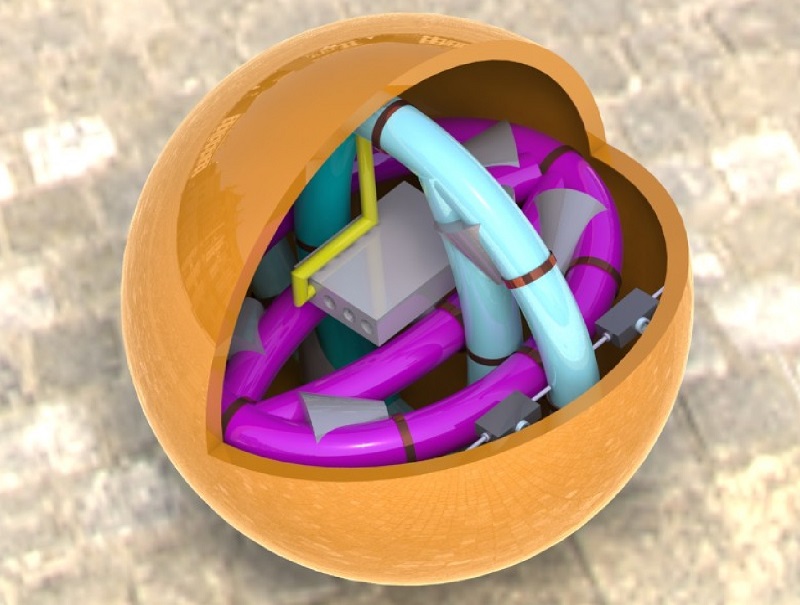}
\caption{"RollRoller", spherical mobile robot.}\label{RollRoller}
\end{figure}

In this paper, we introduce a novel spherical mobile robot named "RollRoller" (see Fig. \ref{RollRoller}). The principle contribution of this paper is validating the 4 degree of freedom unique mechanism (it actuates with angular momentum, gravitational and torque-reaction forces) from derived nonlinear dynamics and numerical simulation. RollRoller moves the core in circular and elliptic pipes in itself to create its motivation force. The required force is produced by circulating the incompressible fluid using hydraulic system. The benefits of this study include:

$\bullet$ The 4 degrees of freedom (e.g., forward/backward, lateral, orientation or jumping abilities) in system is formed parallel with isolated internal driving unit that creates the space for research/rescue tools. We demonstrate our safe hydraulic-base actuator and robot's mechanisms with their characteristics in Section II. 

$\bullet$ The unstable point existing in pendulum-like driving models (gravitational-base and torque-reaction drivers) as major problem is omitted due to the model of RollRoller. The classifications of instability and algorithmic control for forward locomotion are analyzed and designed in Section III.

$\bullet$ Ability to have motion planning by using hybrid nonlinear dynamics and showing how it outperforms its counterpart. In Section IV, we derive the nonlinear dynamics and check the robot about its validity for designed criteria and algorithm performance. 

$\bullet$ Finally, it is demonstrated (see Section V) how this robot with proposed algorithmic approach can have improvements in backward jumping, fluctuations in velocity and linearity of locomotion in Adams/View simulation. Additionally, we have looked to turning action in 3D plain in different frameworks.


\section{Physical Feature}
\subsection{Frame of Robot}
\begin{figure}
\centering
\includegraphics[width=3.3 in]{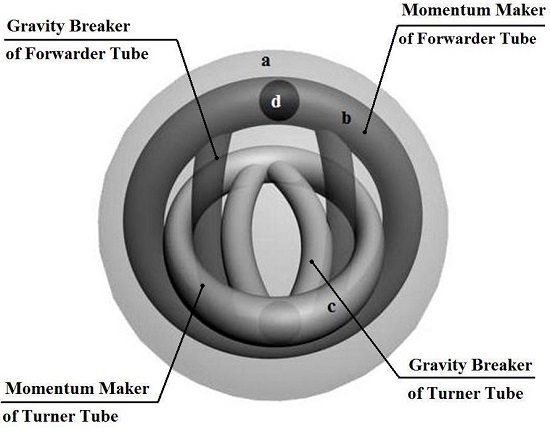}
\caption{a. Spherical Shell, b. Forwarder tube, c. Turner tube. d. Copper   	core.}\label{Fig:FeatureMain}
\end{figure}

RollRoller is covered by plastic glass shell like Fig. \ref{Fig:FeatureMain}-a.
We use pipe-like objects for producing gravity-driven movement force in isolation as a frame and this specification let us to reach flexible mechanism. These tubes are made of polyethylene with high density. Therefore, stamina opposed to unpredicted vibration is neglected and electrical or electromagnetic distortion is impenetrable. In a proposed robot, two groups of these objects are chosen. Main group object is called forwarder tube (FT), placed on striated toward the ground. Next turner tube (TT) is located 90$^o$ tilt to the forwarder tube. For preventing intersection of two objects, the structure of turner tube is made with specific functional change. Accordingly, forwarder tube frame is circular ring. And, turner tube is in elliptic model with certain difference (see Fig. \ref{Fig:FeatureMain}). Combined pipes in each tube have names as their roles in machine. For example, the pipes in shape of half-circle are called momentum maker (MM) lines, however, MM at TT with slightly bend structure (non-circular) is also considered as MM. Gravity breaker (GB) is for the half-elliptic pipes. This configuration difference is based on preventing the intersection of paths of the cores in two tubes (FT,TT). Because the sizes and materials of the leading pipes and also the cores are homogeneous, this mutation is assumed for applying all the required moments from the physical and application point of view. Basically, in both FT and TT, the GBs are inner pipes with shortest distances. We call them MM pipes, because they function in creating the required force to move sphere and conversely GB acts as negative impact rectifier which will be studied in next sections.

For each tube, ball-shaped core with constrained mass $m_c$ is placed.  These cores surfaces are covered with copper layer and filled with high density materials (e.g., tungsten or composite steel). Sphere total mass is $M_s$ without including the cores masses. With $m^*=\frac{m_c}{M_s}$, we have: 
\begin{equation}
\frac{1}{10} \leq m^* <\frac{1}{3} \label{masscoreandsphererelation}
\end{equation}
The reason of constraints in the inequality (\ref{masscoreandsphererelation}) will be discussed in Section IV.B which it gives the core ability to locate itself in the unstable equilibrium points. 
By displacement of these cores inside tubes, the required motion is reachable for the total sphere. 

\subsection{Mechanical Parts}
\subsubsection{Motion Generator by Hydraulic System}
\begin{figure}
\centering
\includegraphics[width=3.3 in]{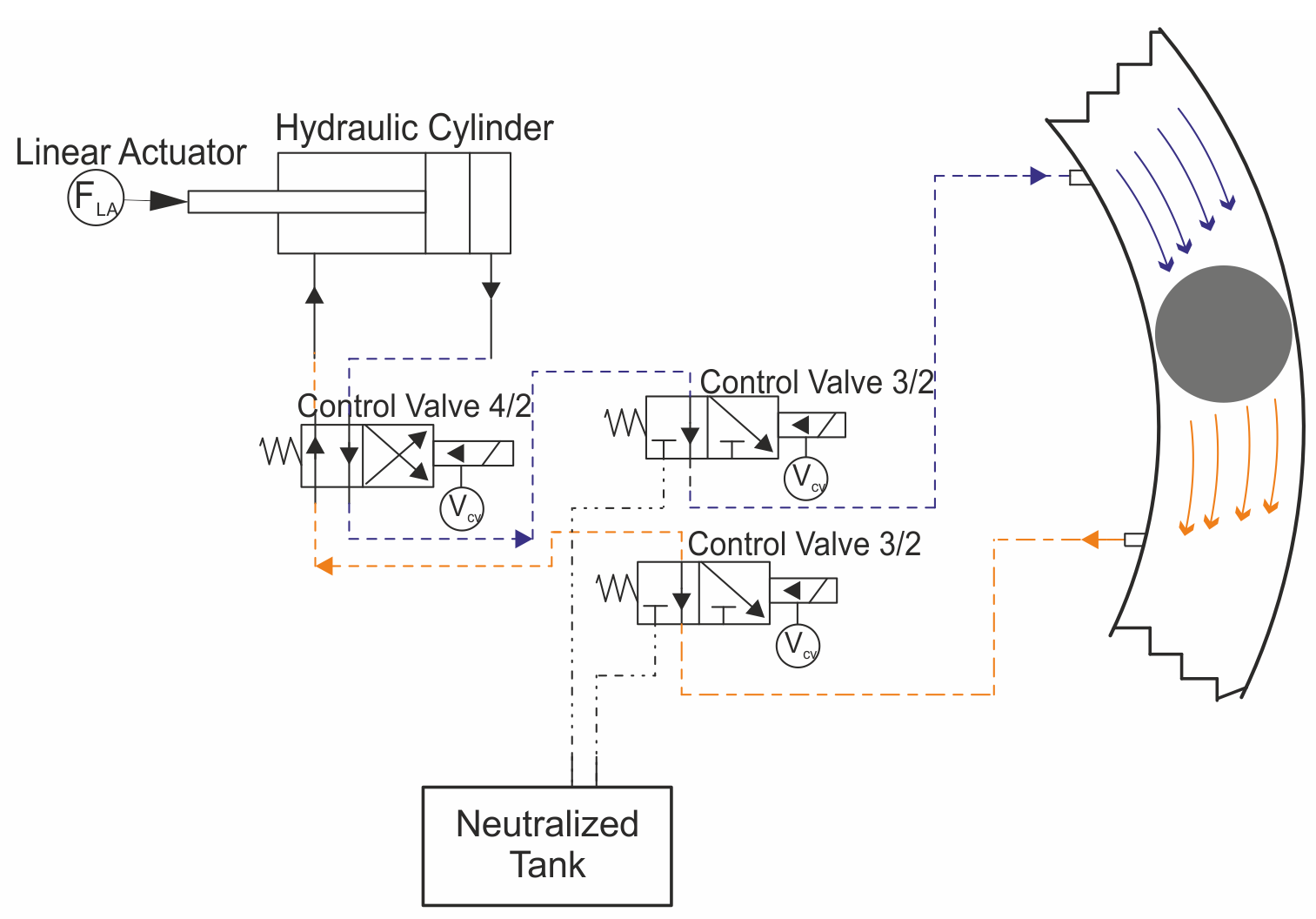}
\caption{Schematic of internal driving unit.}\label{Fig:schamticpneumtic}
\end{figure}

For giving corresponded force to the core, high density flow like superfluid is created by multi-function linear actuators linked to hydraulic cylinder as Fig. \ref{Fig:schamticpneumtic}. 
\begin{figure}
\centering
\includegraphics[width=2.8 in, height=2.4 in]{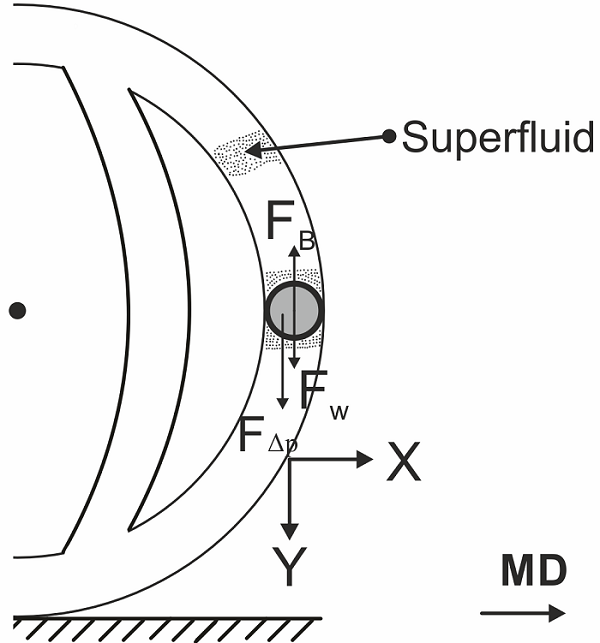}
\caption{External force analysis in the maximum resultant condition.}\label{Fig:Externalforceanalysis}
\end{figure}

\begin{figure}
\centering
\includegraphics[width=1.6 in, height=1.4 in]{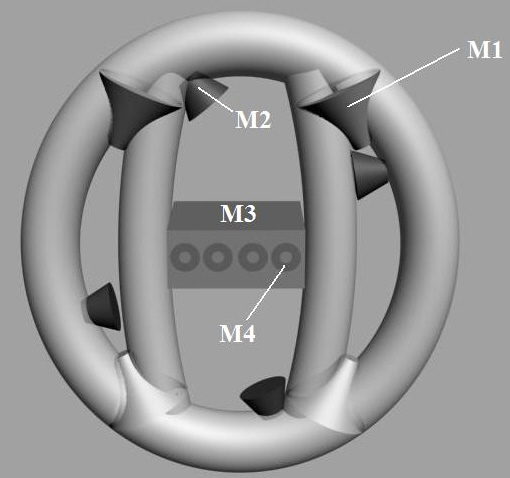}(a)
\includegraphics[width=2.5 in, height=1.4 in]{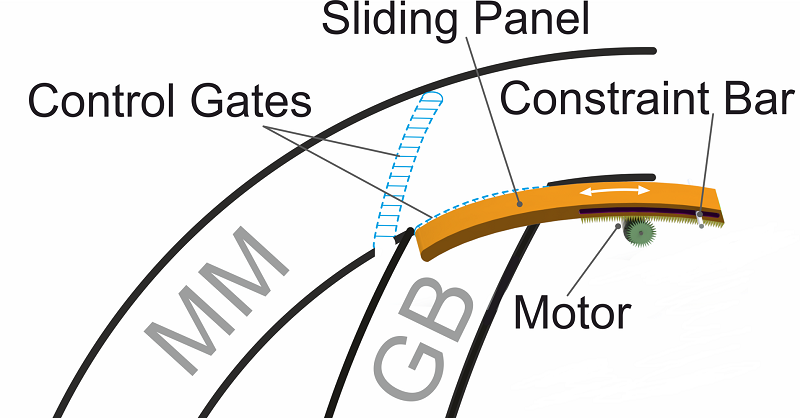}(b)
\caption{(a) Connected mechanical instruments. M1. Control gate module M2. Input hole for fluid entrance M3. Motor box M4.Output of fluid, (b) Sliding control gates.}\label{Fig:mechanicandgate}
\end{figure}

In this system, hydraulics will not be used as actuator instead they are the impulsive fluid replacer without requiring pressurized tank, compressor and pressure regulator inside internal driving unit. The reason is that the total features and internal driving unit mechanism are placed in closed system. The flow is circulated through pipes, hydraulic cylinders and neutralized tank. In this system neutralized tank is responsible to keep fluid in itself and stop it to enter a cylinder in certain times. This happens when fluid flow has certain speed and it may waste energy or oppose the hydraulic actuator task to enforce linear actuator unwillingly. To understand efficiency and required input force to move the core, we assume that core is in a position with maximum gravitational effect $F_w$ and there is certain available static friction between core and pipe $F_f$ (see Fig. \ref{Fig:Externalforceanalysis}). The $F_B$ and $F_{\vartriangle P}$ stands for buoyancy force and created force due to pressure difference of fluid in core's upper and lower bond. In y direction we have: 
\begin{equation}
F_{c_{Max}}=F_w+F_{\vartriangle P}-F_B-F_f \label{maximumgravity}
\end{equation}
Where $m_{lc}$ and $m_b$ are the masses of the core and displacing fluid. Taking into consideration if superfluid is used inside pipes, the viscosity tends to the negligible value hence $F_f$ is ignored.
\begin{multline}
F_{c_{Max}} =F_w+F_{\vartriangle P}-F_B=(m_{lc}-m_b)g+\triangle P A_c =m_cg+\triangle P A_c
\label{lastbyonecee} 
\end{multline}
Where $m_c$ is apparent mass of core and $A_c$ is the projection of core area which is perpendicular to y direction. $\triangle P$ tells the efficient pressure difference of the bottom and top of the core inside the pipe.
From Eq.(\ref{LAtoCore}) the transmitted force from linear actuator to core under the condition that the fluid is in incompressible condition is obtained. The pressure inside the system is considered to be enough to have our operation in perfect manner. The pure enforced motivation toward core is created from sum of both $P_U = \frac{4 F_{LA}}{\pi (D^2_1-D^2_2)}$ and $P_L=\frac{4 F_{LA}}{\pi D^2_1}$ as upper and lower chamber created pressure from hydraulic cylinder connect to linear actuator (i.e. $F_{LA}$ is input force of linear actuator). The $D_1$ and $D_2$ stand for the full bore piston diameter and piston rod diameter of 
cylinder, respectively.The $r_c$ is the radius of the core. Due to both sides support in locomotion of core, the pressures are summed to result the total active force as Eq. (\ref{LAtoCore}).
\begin{equation}
F_{c_{T}}= 8r_c^2F_{LA}[\frac{2D^2_1-D^2_2}{D^2_1(D^2_1-D^2_2)}]
\label{LAtoCore}
\end{equation}
As a result, to have the core in movable condition with certain accuracy and efficiency, Eq. (\ref{ComprassionLAandCore}) has to be satisfied. Moreover, Eq. (\ref{ComprassionLAandCore}) stays applicable when the leakage from one side of the core to other side is insignificantly small (tangent surfaces between sphere-core).
\begin{equation}
F_{c_{T}} > F_{c_{Max}}
\label{ComprassionLAandCore}
\end{equation}
This produced fluid force enters the tubes with steady magnitude under ideal condition. However, main goal is to keep cores in certain speed hence fluid actuated force can be vary manually. In Fig. \ref{Fig:mechanicandgate} (a) the internal driving unit is placed inside a box and output ports are located on the box in the robot. These systems can perform in either suction or injection function. For each forwarder tube and turner tube, four motors separately are placed in motor boxes. Each of boxes contains four manual input holes connected to specific places on tube as seen in picture. This overall system has got a capability of controlling the core in most variety of control algorithms.     
\subsubsection{Control Gates}
Control gate modules are located in accurate places around the tube for controlling path of the core via opening and closing. Fig. \ref{Fig:mechanicandgate} (a)-M1 shows that, this instrument is placed as interface of MM with GB pipes. For practical use of each schematic gate modules (symbolized cones in Fig. \ref{Fig:mechanicandgate} (a)-M1), two sliding control gates are placed as Fig. \ref{Fig:mechanicandgate} (b). The sliding gates act to open or close related pipe passage beside keeping the core to follow the smooth and curvy locomotion while passing through. As a functioning stages for gate modules, during function I activation, it is assumed both pipes are closed so any fluid movement and possibility of entering the core to areas are avoided. Function II and III keep an either path (e.g., MM open and GB close) open or closed. Motion of the core will depend on the function mode of gates, created fluid displacement and control valves mode. These choices to displace robot via integration of different mechanical parts, give freedom to system to have more control over itself without constraining its locomotive degrees of freedom. 
\section{Movement Analysis}
\subsection{Analytical Criteria}
\begin{figure}
\centering
\includegraphics[width=3.2 in]{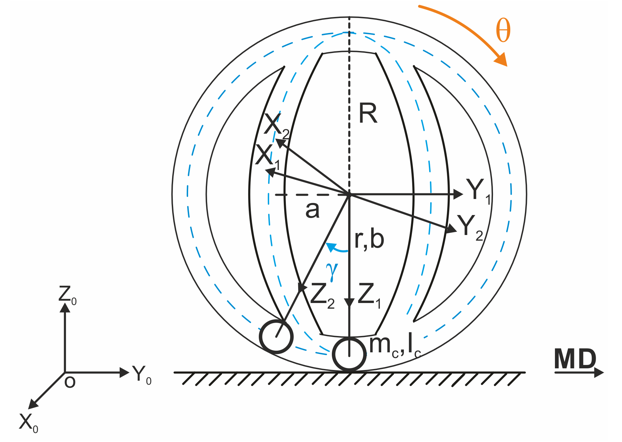}
\caption{Rolling motion model along O-y.}\label{Fig:SphereGeometryCore}
\end{figure}

Most of pendulum driven and mass displacement mechanisms in SMRs make them impracticable for internal driving units to bring the center of mass to above upper-half of the sphere. That phenomenon is avoided in this mechanism via Eq. (\ref{masscoreandsphererelation}), however, this instability bring three different behaviors relevant to position of the core, speed of the core and sphere. To analyze these morphologies, it is assumed that TT's core is in middle of GB so there will be no effect on diversity of center of mass related to total robot.  \\
The graphical notations for side view of "RollRoller" is  as Fig. \ref{Fig:SphereGeometryCore}. 
Reference frames are denoted in which $O_0-X_0Y_0Z_0$ represents the inertial fixed reference. The moving frame connected to the center of sphere is $O_1-X_1Y_1Z_1$, which
translates only with respect to inertial fixed reference. Also, $O_2-X_2Y_2Z_2$ is another
frame attached to center like previous one but it rotates only with respect to the
$O_1-X_1Y_1Z_1$. Table \ref{variabletable} shows the defined variables for rolling spherical robot. 
\begin{table}[ht!]
\renewcommand{\arraystretch}{1.3}
\caption{Nomenclature of dynamic analysis}
\label{variabletable}


\begin{tabular*}{\textwidth}{c @{\extracolsep{\fill}} ccccc}
\hline
$\theta$ & Rolling angle of RollRoller around x axis\\
$\gamma$ & Locomotion angle of the core\\
g & Gravitational acceleration\\
R & Sphere radius\\
r & Distance between the center of sphere and core\\
a & Semi-minor axis of GB\\
b & Semi-major axis of  GB\\
\hline
\end{tabular*}
\end{table}
We are considering that the core is traveling through the dashed line. Also, MD basically stands for main direction of "RollRoller". The core distance (r) is varied between the semi-minor (a) and semi-major (b) spaces. It is assumed that there is no slip between sphere and ground. The core unstable region is $-\frac{\pi}{2} <\gamma< \frac{\pi}{2}$, the location of $- \pi $ is assumed the stable point and $^{+}_{-}\frac{\pi}{2}$ location looked as stable and maximum gravitational points for sphere motion. Due to similarity of $-\pi <\gamma< -\frac{\pi}{2}$ and $\frac{\pi}{2} <\gamma< \pi$, we choose the $-\pi <\gamma< -\frac{\pi}{2}$ region for our analysis. Each case gives brief summary of the core and sphere locomotion with diverse state conditions in the instability area. 
\begin{figure}[t!]
\centering
\includegraphics[width=1.8 in]{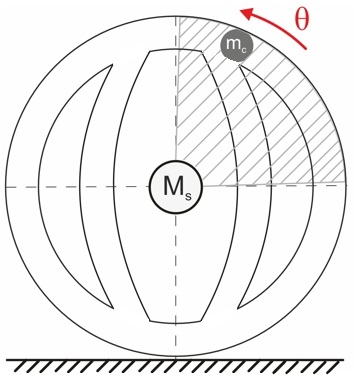}(1)
\includegraphics[width=1.8 in]{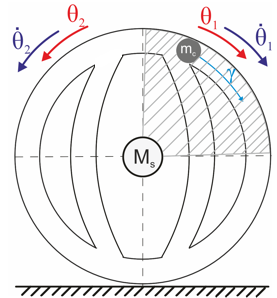}(2)
\includegraphics[width=1.8 in]{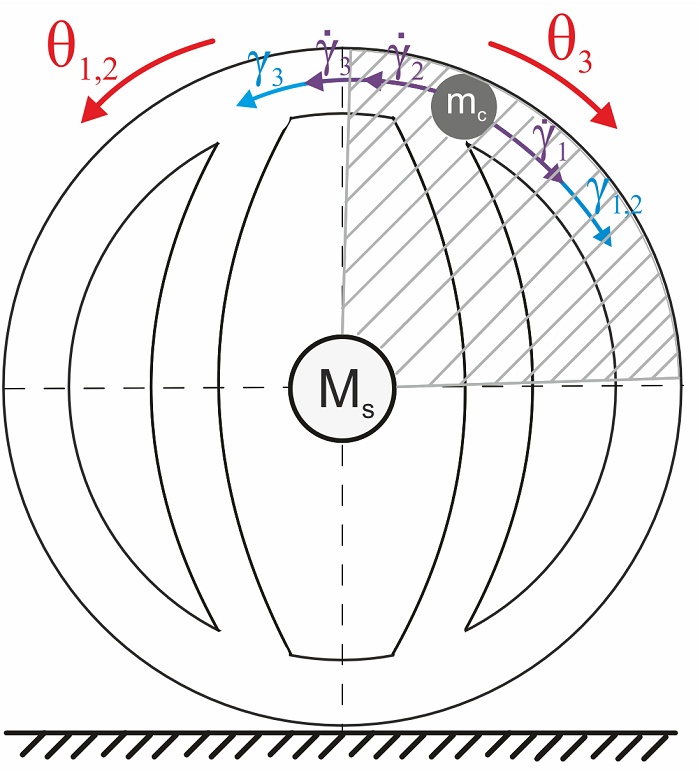}(3)
\includegraphics[width=1.9 in]{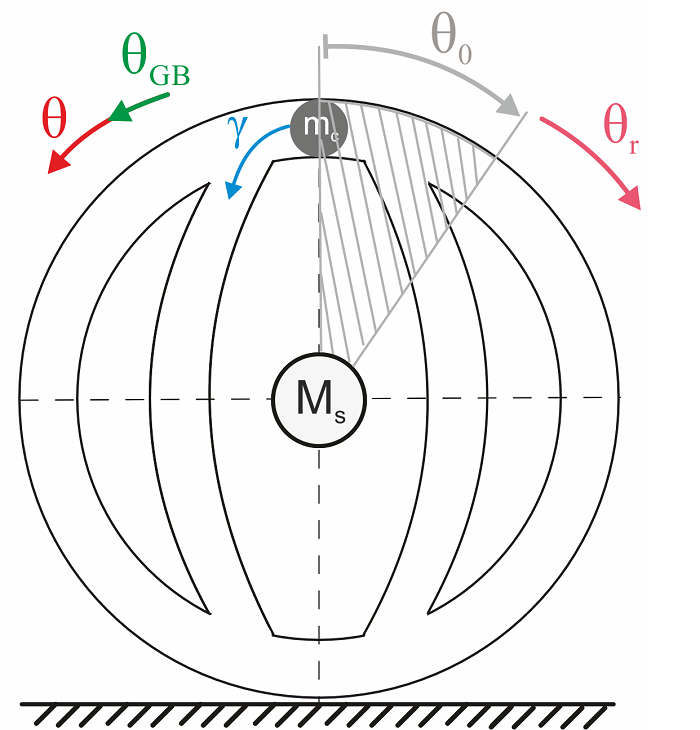}(4)
\caption{(1) The sphere's motion with no initial velocity,(2) The sphere's motion with initial sphere velocity,(3) The sphere's motion with initial core velocity,(4) GB pipe core and sphere motion in "RollRoller".}
\label{Fig:3formoflocomotion}
\end{figure}
The core motion through MM:

\textbf{Case 1:}Neither the sphere nor the core has initial velocity so this results to create locomotion for sphere in opposite direction of MD as Fig. \ref{Fig:3formoflocomotion}-1 (Initial Conditions: $\theta = 0\;,\; \dot{\theta} = 0\;,\; -\pi <\gamma< -\frac{\pi}{2}\;,\; \dot{\gamma}=0$).

\textbf{Case 2:}The sphere with initial velocity , leads the robot through initial response with effect of the core location. However, with given constraints, sphere follows initial velocity direction (Initial Conditions: $\theta = 0\;,\;\dot{\theta}_{min} <|\dot{\theta}|<\dot{\theta}_{max}\;,\; -\pi <\gamma< -\frac{\pi}{2}\;,\; \dot{\gamma}=0$). For certain level, the sphere velocity doesn't create enough kinetic energy.

\textbf{Case 3:} By $\dot{{\gamma}}_1>0$ the total sphere moves counterclockwise direction with having clockwise threshold value ($\dot{\gamma}_{CT}$) for core velocity (Initial Conditions: $\theta_1 = 0\;,\;\dot{\theta}_1=0\;,\; -\pi <\gamma_1< -\frac{\pi}{2}\;,\; 0<\dot{\gamma}_1<\dot{\gamma}_{CT}$). When core velocity is $\dot{{\gamma}}_2<0$ till counterclockwise threshold velocity 
$\dot{\gamma}_{CCT}$, sphere follows the counterclockwise movement (Initial Conditions: $\theta_2 = 0\;,\;\dot{\theta}_2=0\;,\; -\pi <\gamma_2< -\frac{\pi}{2}\;,\; \dot{\gamma}_{CCT}<\dot{\gamma}_2<0$). And, It is expected to give opposite responses  for $\dot{{\gamma}}_{1,2}$, while they are exceeding the threshold velocity. $\dot{\gamma}_{3}$ as an example that surpass  the value from $\dot{\gamma}_{CCT}$ and results robot moves in clockwise direction (Initial Conditions: $\theta_3 = 0\;,\;\dot{\theta}_3=0\;,\; -\pi <\gamma_3< -\frac{\pi}{2}\;,\; \dot{\gamma}_3<\dot{\gamma}_{CCT}$).

Results in section IV will show how RollRoller covers these logical hypothesis to omit the instability points. Additionally, Case 1 is true under the condition of free fall mode for the core. Therefore, either if we fix the core in its location or it has high friction between the core and pipe surfaces, total sphere will pull the core to neutralization point under $M_s$ location. As a result, $\theta$ will turn clockwise.

Fig. \ref{Fig:3formoflocomotion}-4 gives the sight to advantages of GB in "RollRoller", which will have direct impact on enhancing locomotion with including more freedom in robot's motion. The obtained results are shown in section IV.B.

\textbf{Case 4:} The core pass through the GB in the sphere with initial angle of $\theta_0$. Resulted $\gamma$ moves the sphere in counterclockwise direction with $\theta_{GB}$ which $\theta_{GB}<<\theta$. $\theta$ in this case stands for locomotion of sphere if the core moves through MM (Initial Conditions: $0\leq \theta \leq \theta_{T},\dot{\theta}=0, \gamma = k\pi, 0\leq \dot{\gamma}$). $\theta_{T}$ is the boundary for sphere's slope in the unstable region. In the same way, if the total sphere has motion in clockwise direction which is in opposition with the core motion, the latest displacement $\theta_r$ after reaching the ground equilibrium point will be as Eq. (\ref{Eq:Case4GBmotion}) for both MM and GB cases.
\begin{equation}
\theta_r-\theta_{MM}<\theta_r-\theta_{GB}
\label{Eq:Case4GBmotion}
\end{equation}

\subsection{Forward Direction}
\begin{figure}
\centering
\includegraphics[width=2 in]{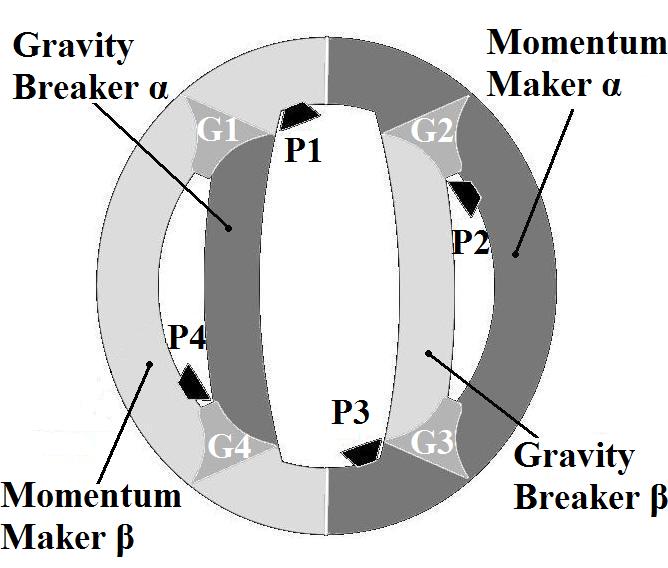}
\caption{Side way view and partition definition.}\label{Fig:sideviewexplanationFD}
\end{figure}
\begin{figure}
\centering
\includegraphics[width=4.8 in]{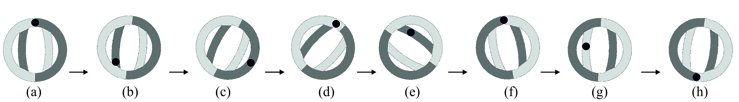}
\caption{The cycles of forward locomotion.}\label{Fig:forwarddirectionmotion}
\end{figure}
The forward direction motion of robot is studied with including the gates, input ports and pipeline paths (i.e., MM and GB). As shown in Fig. \ref{Fig:sideviewexplanationFD}, overall tube is divided to subparts on FT. "$\alpha$" region pipes are the initiation step for locomotion and next cycle is taken by "$\beta$" region. $Pn$ and $Gn$ in symbolic way stand for function stages of fluid flow input/output ports and control gates, respectively.

The cylinder input and output ports are connected in series. For instance, first cylinder's two output ports are connected to P1 and P2. The gates are stated between the bi-fraction of hydraulic cylinders (e.g. $Hyd._1 \rightarrow P1-P2$ and $Hyd._3 \rightarrow P3-P4$).

The proposed steps are created in ideal locomotion for robot without any slippery. The relative basic control law, Algorithm \ref{Algo:forwarddirection}, for perfect locomotion (Explanation in Section V) is created.
The cycles in Fig. \ref{Fig:forwarddirectionmotion} are illustrated as a sample for all of possible situations. 
Each cycle is independent from others so the robot can stay in the specific rotation in great deal of time (i.e. Fig. \ref{Fig:forwarddirectionmotion}-c) till reaches to the relatively similar form in other step.
\alglanguage{pseudocode}
\begin{algorithm}
\caption{Forward Locomotion}
\begin{algorithmic}[1]
\Procedure{Right Direction Motion}{$\theta,\dot{\theta},\gamma,\dot{\gamma}$}
\State Let $\gamma$'s location determine the region X
\Comment{X is $ \alpha / \beta$}
\While{$0<\dot{\gamma}\leq\dot{\theta}$} 
\State Keep the core moving in MM of region X
\EndWhile
\While{$\dot{\theta}<\dot{\gamma}$}
\If {$-k\pi-\frac{\pi}{\eta_{\gamma}}<\gamma \leq -k \pi+\frac{\pi}{\eta_{\gamma}}$ and $k\pi+\frac{\eta_{\theta}\pi-\pi}{\eta_{\theta}} \leq \theta \leq k\pi+\frac{\eta_{\theta}\pi+\pi}{\eta_{\theta}}$}
\State Pass the core from GB of region X
\Else
\State Keep core moving in MM of region X
\If {$(2k+1) \pi \theta \approx \gamma$ }
\State Pass the core through the GB of region X
\State Switch the region X 
\State \textbf{return} $3$
\EndIf
\EndIf
\EndWhile
\EndProcedure
\end{algorithmic}
\label{Algo:forwarddirection}
\end{algorithm}

To implement the hybrid locomotion with GB and MM interactions Algorithm \ref{Algo:forwarddirection} is designed. In the Algorithm \ref{Algo:forwarddirection}, first while loop represents the time that speed of the core is equal or less than speed of the sphere (see Fig. \ref{Fig:forwarddirectionmotion}-c), this behavior is considered as torque reactive form of motion which is interpreted as the pendulum driven models. 
Next while loop is for the special cycles that core moves from GB (see Fig. \ref{Fig:forwarddirectionmotion}-d ,e and g), so the core accelerates faster and consequently causes increasing the momentum of inertia. In line 7, the $\eta_{\theta}$ and $\eta_{\gamma}$ parameters are the regional constraints related to activation location for GB in which it is a main design function in Algorithm \ref{Algo:forwarddirection}. Impact of this particular constraint will be evaluated in Section V on RollRoller locomotion. This is the part that system works with torque-reaction and angular momentum methods. 
Also, we can have only gravity-base motion by moving the core considerably fast and lock it in specific location in MM/GB pipes. 
It is believed, this is the first mechanism that can have all three motivation forces uniquely.
As an example for the integration of control gates and input ports with the cycle of locomotion in Fig. \ref{Fig:forwarddirectionmotion} for right direction motion, the Table $\ref{tab:forwarddirection}$ is obtained.

\begin{table}[ht!]

\renewcommand{\arraystretch}{1.3}
\caption{Functions of forward movement}
\centering

\begin{center}
\begin{tabular*}{\textwidth}{ccccccccc @{\extracolsep{\fill}} ccccc}
\hline 
Cycle,  & \multicolumn{1}{c}{} & \multicolumn{1}{c}{Gate} & \multicolumn{1}{c}{Modules} &  & \multicolumn{1}{c}{} & \multicolumn{1}{c}{Output} & \multicolumn{1}{c}{Ports} & \tabularnewline
\cline{2-9} 
Core Loc. & G1 & G2 & G3 & G4 & P1 & P2 & P3 & P4\tabularnewline
\hline 
a, EP& II & II & I & II & O & IT & ST & O\tabularnewline

b, GB $\alpha$& II & I & I & II & IT & O & ST & O\tabularnewline

c, MM $\alpha$& I & II & II & I & O & ST & IT & O\tabularnewline
 
d, MM $\beta$ & II & I & I & II & IT & O & ST & O\tabularnewline

e, GB $\alpha$ & II & I & II & II & IT & ST & ST & O\tabularnewline
 
f, MM $\beta$ & I & II & III & III & ST & O & O & IT\tabularnewline

g, GB $\beta$ & I & III & III & I & ST & O & IT & O\tabularnewline

h, EP & III & II & I & I & O & IT & O & ST\tabularnewline
\hline
\end{tabular*}
\end{center}
\centering EP = Equilibrium Point, ST = Suction in Tube, IT = Injection in Tube, O = Off Mode, I = Closed Gate, II = Opened $\alpha$ and closed $\beta$, III = Opened $\beta$ and closed $\alpha$.
\label{tab:forwarddirection}
\end{table}
\subsection{Turning Motion}
To have clear understanding about RollRoller's morphology on 3D plain, the turning ability of robot was studied as one of imperative degrees of freedom. 

\begin{figure}
\centering
\includegraphics[width=3.9 in]{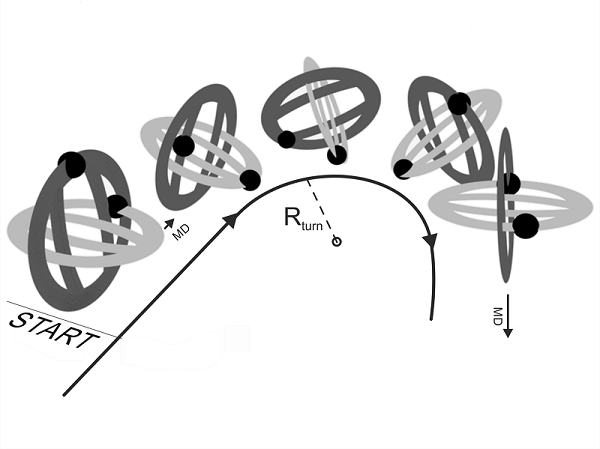}
\caption{Schematic of turning motion.}\label{Fig:CircularTurningSchematic}
\end{figure}
The TT core's mass is assumed with small fraction of increase $\delta$ due to deformed structure of TT (property in Eq. (\ref{masscoreandsphererelation})). $\delta$ is approximated depending on the structure of TT and material of the sphere. The proposed locomotion manner for turning is developed as schematic in Fig. \ref{Fig:CircularTurningSchematic}. 
\begin{equation}
\frac{M_s}{10}+\delta \leq m_{c_{TT}}<\frac{M_s }{3}+\delta 
\label{Eq:TTMASsPROPERTy}
\end{equation}

In this form of locomotion, while FT is following the cycles in forward motion (see Fig. \ref{Fig:forwarddirectionmotion}) with core velocity that tending to exceeds the sphere velocity, the TT core tend to move from MM to corner equilibrium point (intersection point of GBs and MMs) on right hand side based on the MD. By maintaining the TT core in this location through the motion period, we expect the robot to have turning with $R_{turn}$ radius. In section V.B, the turning maneuver will be shown and compared on 3D plain with Adams/View simulation.
\section{Dynamics}
\subsection{Forward Direction Nonlinear Dynamics}
\begin{figure}
\centering
\includegraphics[width=3 in]{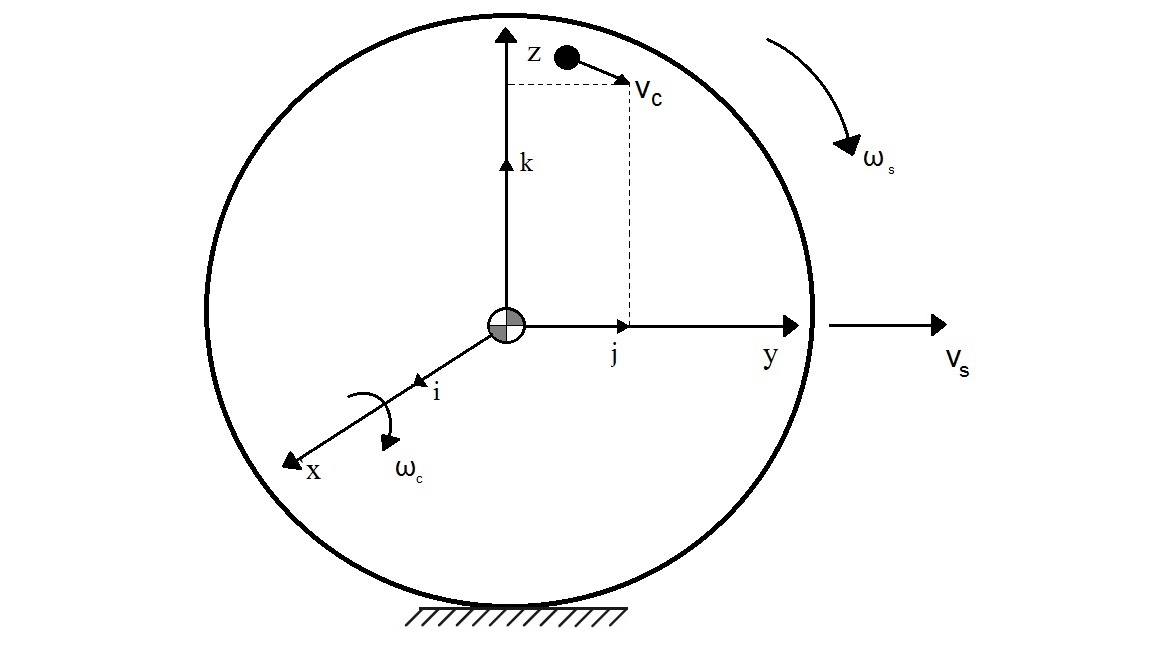}
\caption{The unit vectors, angular and linear velocities.}
\label{Fig:Unitvectorssphere}
\end{figure}

To derive the dynamics of the system, we used variables and geometry models from section III, and Fig. \ref{Fig:Unitvectorssphere} shows the basic coordination and units of our operation.

$D_{po}$  represents the position vector in the robot.  Angular and linear velocity vectors are $\omega_s$ and $V_s $ for total sphere, also angular velocity $\omega_c$ and linear velocity $V_c $  are related to the core.
\begin{equation}
D_{po}=-a\sin (\gamma+\theta) \hat{j}-b \cos(\gamma+\theta) \hat{k}
\end{equation}
\begin{equation}
\omega_s=\dot{\theta}\hat{i}
\end{equation}
\begin{equation}
V_s=R\dot{\theta}\hat{j}
\end{equation}
\begin{equation}
\omega_c=(\dot{\gamma}+\dot{\theta})\hat{i}
\end{equation}
\begin{equation}
V_c=(R\dot{\theta}+a(\dot{\gamma}+\dot{\theta}) \cos(\gamma+\theta))\hat{j}+(b(\dot{\gamma}+\dot{\theta})\sin(\gamma+\theta))\hat{k}
\end{equation}

For obtaining the dynamic model of robot, the Lagrangian function L containing only the terms due to rotation along the O-y axis, as follows: 
\begin{equation}
L=E_k-E_p \label{Eq:kineticpotentialenergy}
\end{equation}
In Eq. (\ref{Eq:kineticpotentialenergy}), $E_k$  and $E_p$ stand for total kinetic energy and potential energy. $I_s$ and $I_c$  show the masses' moment of inertia for the sphere and the core.
\begin{equation}
L=\dfrac{1}{2}M_s|V_s|^2+\dfrac{1}{2}I_s|\omega_s|^2+\dfrac{1}{2}m_c|V_c|^2+\dfrac{1}{2}I_c|\omega|^2\\
+m_cgd_{c-z}
\end{equation}
Next, we just substitute parameters in Lagrangian function. This equation is mutated generally for both elliptic and circular motion as Eq. (\ref{Eq:lastlagrangian}).  
\begin{multline}
L=\dfrac{1}{2}R^2\dot \theta^2M_{s}+\dfrac{1}{2}I_s\dot \theta^2+\dfrac{1}{2}I_c(\dot\gamma+\dot \theta)^2
\\
+\dfrac{1}{2}m_c[(R\dot{\theta}+a(\dot{\gamma}+\dot{\theta})\cos(\gamma+\theta))^2+(b(\dot{\gamma}+\dot{\theta})\sin(\gamma+\theta))^2]\\
+m_cg\frac{ab}{\sqrt{(a\cos(\gamma+\theta))^2+(b\sin(\gamma+\theta))^2}}\cos(\gamma+\theta) 
\label{Eq:lastlagrangian}
\end{multline}
The gravitational potential energy is adapted to have a true mode when the core is following either elliptic or circular lines in RollRoller. We assume there are distinct viscous frictions for core and sphere surfaces. The general energy dissipation function is written for Lagrangian with dependency on velocities of model. 
\begin{equation}
P=\frac{1}{2} \sum \zeta_i \dot{q}_{i}^{2}=\frac{1}{2}(\zeta_{\theta} \dot{\theta}^2 + \zeta_{\gamma} \dot{\gamma}^2)
\end{equation}
The Lagrange-Euler equations for translation in O-y plain is as follows:
\begin{equation}
\dfrac{d}{dt}(\dfrac{\partial L}{\partial \dot q_i})- \dfrac{\partial L}{\partial q_i}+\dfrac{\partial P}{\partial \dot{q}_i}= \tau_i
\end{equation}
Where our $q_i$ are $\theta$ and $\gamma$, respectively. Also, the input torque directly applied to the core $\tau_{\gamma}=bsF_{c_{T}}$ (s is the number of involved hydraulics) and there is no supportive/reactive force toward the total sphere $\tau_{\theta}=0$.

Therefore, equations of motion for mechanical system are as :
\begin{multline}
\left[\begin{array}{ccc}
M_{11} & M_{12} \\
M_{21} & M_{22}\\
\end{array}\right]
\left[\begin{array}{ccc}
\ddot{\theta} \\
\ddot{\gamma}\\
\end{array}\right]+\left[\begin{array}{ccc} N_{11}\\ N_{21}\\ \end{array}\right]+\left[\begin{array}{ccc} G_{11}\\ G_{21}\\ \end{array}\right]
=\left[\begin{array}{ccc} \tau_\gamma \\ \tau_\theta \\ \end{array}\right]\space\space\space\space\space\space\space\space
\label{Eq:motionmechanismsystem}
\end{multline}
Where parameters are defined by $M_{ij}$ the acceleration coefficients , $N_{ij}$ velocity dependencies and $G_{ij}$ gravity factors, while:
\begin{multline}
\begin{split}
M_{11}&=I_c+a^2m_c-aRm_c\cos(\gamma+\theta)+m_c\sin^2(\gamma+\theta)(b^2-a^2)  \\
M_{12}&=I_c+a^2m_c+m_c\sin^2(\gamma+\theta)(b^2-a^2) \\
M_{21}&=M_sR^2+I_s+m_cR^2-2aRm_c\cos(\gamma+\theta)+Ic
\\
&+a^2m_c+m_c\sin^2(\gamma+\theta)(b^2-a^2) \\
M_{22}&=I_c+a^2m_c+m_c\sin^2(\gamma+\theta)(b^2-a^2)-aRm_c\cos(\gamma+\theta)\\
N_{11}&= -\dot{\theta}^2[aRm_c\sin(\gamma+\theta)] \\
&-(\dot{\gamma}+\dot{\theta})^2[m_c\sin(\gamma+\theta)\cos(\gamma+\theta)(b^2-a^2)]\\
&-(\dot{\gamma}\dot{\theta})(aRm_c\sin(\gamma+\theta))+\zeta_{\gamma} \dot{\gamma}\\
N_{21}& = -\dot{\theta}^2[aRm_c\sin(\gamma+\theta)]-(\dot{\gamma}+\dot{\theta})^2[m_c\sin(\gamma+\theta)\cos(\gamma+\theta)(b^2-a^2)]\\
&-(\dot{\gamma}\dot{\theta})(aRm_c\sin(\gamma+\theta))+\zeta_{\theta} \dot{\theta}\\
G_{11}&= G_{21} = abm_cg[\frac{\sin(\gamma+\theta)(a^2\cos^2(\gamma+\theta) }{...}\\
&\frac{+b^2\sin^2(\gamma+\theta))+\frac{\sin(\gamma+\theta)\cos^2(\gamma+\theta)(b^2-a^2)}{a^2\cos^2(\gamma+\theta)+b^2\sin^2(\gamma+\theta)}}{a^2\cos^2(\gamma+\theta)+b^2\sin^2(\gamma+\theta)}] 
\end{split}
\label{Eq:NONLINEARDynamics}
\end{multline}
We orientate the nonlinear dynamics in this subsection in Eq. (\ref{Eq:motionmechanismsystem}) as second order equations to help us to have better view over the system.

\subsection{Dynamics Results}
\begin{figure}[h!]
\centering
\includegraphics[width=4.8 in, height=2.4 in]{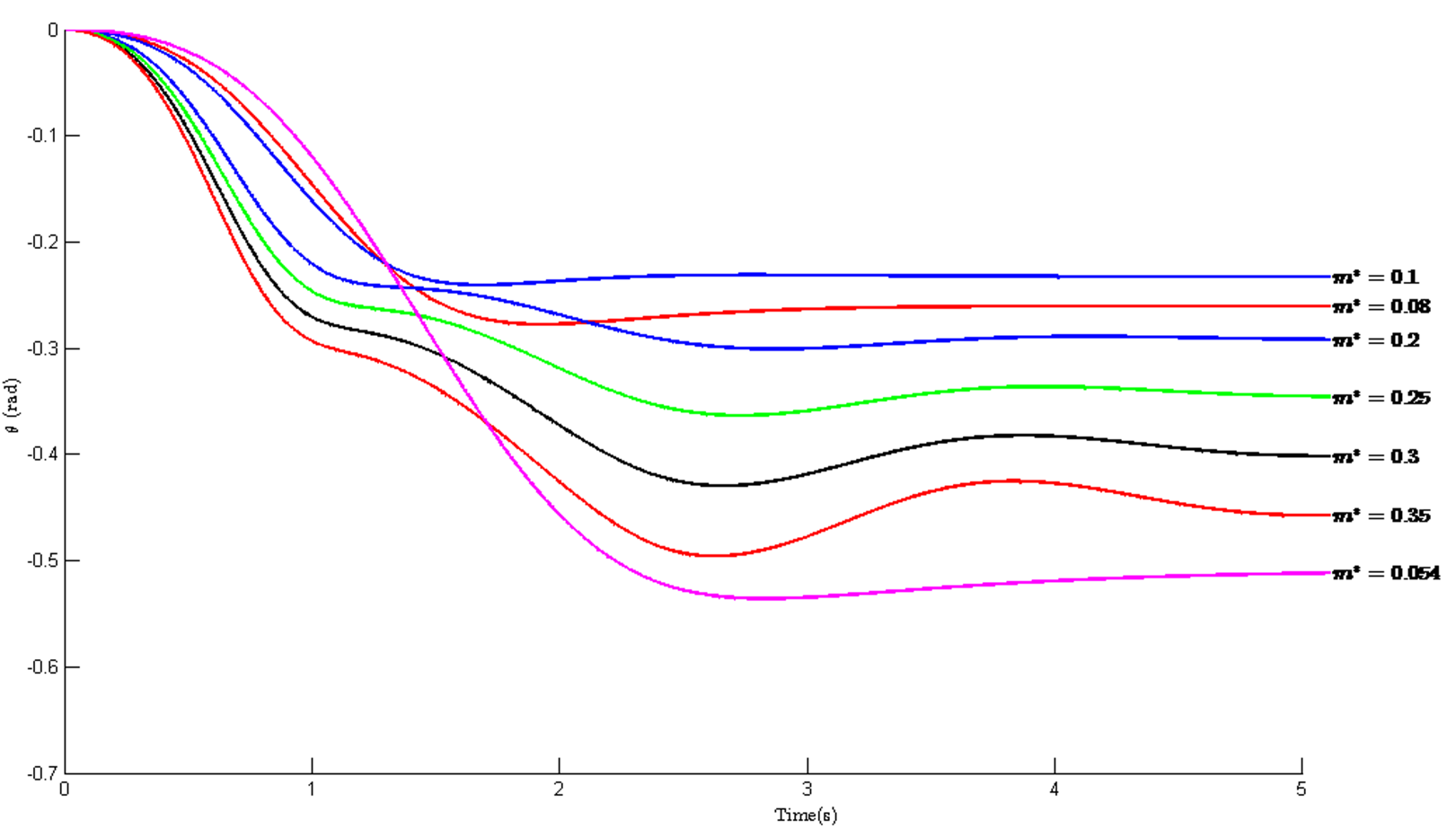}
\caption{Instability behavior in motion of sphere for different mass ratios.}\label{Fig:instabilityanalysisfordifferentmass}
\end{figure}
We analyze the dynamics for two basic points; first, about verifying the proposed criteria in section  III for omitting the instability point. Second, to show the efficiency of proposed algorithm. The derived nonlinear dynamics in Eqs. (\ref{Eq:motionmechanismsystem})-(\ref{Eq:NONLINEARDynamics}) are solved via ODE45 solver \cite{DORMAND1980ODE,Matlab1997ODE} in Matlab. We used variables from Table \ref{Tab:NumericalVariabledyn} in evaluating the motion of system. 

\begin{table}
\renewcommand{\arraystretch}{1.3}
\caption{Data of RollRoller's dynamic variables}
\label{Tab:NumericalVariabledyn}
\centering
\begin{tabular*}{\textwidth}{@{\extracolsep{\fill}}cccc}
\hline
Variable & Value & Variable & Value\\
\hline
$g$ & $9.8 \;\; m/s^2$ & $b$ & $0.131 \;\;m$\\
$M_s$ &$ 1 \;\;kg$ & $a$ & $0.0450\;\;m$\\
$m_c$ & $0.25\;\; kg$ & $I_{c_{MM}}$ & $0.0402\;\;kg.m^2$\\
$\zeta_{\theta} $ & $0.2$ & $I_{c_{GB}}$ & $7.7440\times10^{-4}\;\;kg.m^2$\\
$\zeta_{\gamma} $ & $<0.1$ & $I_s$ &  $0.0140\;\;kg.m^2$ \\
$R$ & $0.145\;\;m$&&\\
\hline
\end{tabular*}
\end{table}
The inertia coefficients of sphere and core in MM and GB pipes are: 
\begin{equation}
\begin{split}
I_s&=\frac{2}{3}M_sR^2\\
I_{c_{MM}}&=\frac{2}{5}m_cb^2\\
I_{c_{GB}}&=\frac{2}{5}m_c(\frac{a+b}{2})^2\\
\end{split}
\label{Eq:Inertia}
\end{equation}
The average of semi-minor and semi-major distance has been substituted in order to estimate and settle the momentum of inertia in the system to its true value when entering to GB path. The viscous friction of sphere is adapted to the environment that is approximately on an even terrain. The core displacement also has the minimum dissipation energy by assuming the frictionless surface between core and interior part of pipe. 
\subsubsection{$m^*$ Range Validation}
As it is seen from inequality (\ref{masscoreandsphererelation}), to obtain the applicable domain of mass ratio for sphere and core, we analyze the dynamics in $X_0=[\theta \; \; \dot{\theta}\;  \; \gamma \; \;\dot{\gamma}]=[0\; \; 0\; \;  -\frac{3\pi}{2}\; \;  0]$ initial conditions with no input torque. Also the dissipation coefficients for core and sphere are assumed $\zeta_\gamma= 0.01$ and $\zeta_\theta= 0.2$. As it is seen from Fig. \ref{Fig:instabilityanalysisfordifferentmass}, the stability domain is limited between $0.1$ - $0.3$. For values over lower and upper limits system follows abnormal displacement along the axis. Also, due to considerable high amplitude ripples after 0.3 mass ratio the results become uncontrollable. The chosen value for mass of the core and $m*$ ($m_c=0.25$ kg and $m*=0.25$) in Table \ref{Tab:NumericalVariabledyn} is the value with minimum overshoots (nearly $5 \%$) and settling time ($5.2$ sec) which are the characteristics of under-damped stable system for this study.
\begin{figure}
\centering
\includegraphics[width=4.8 in, height=3.4 in]{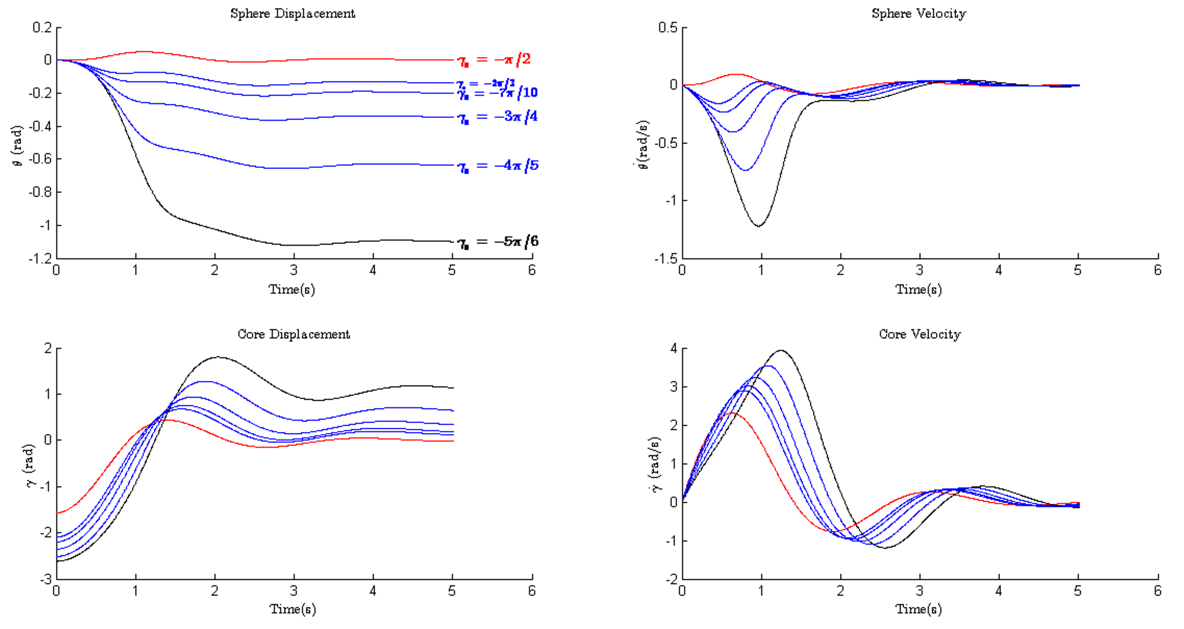}\newline
(a)
\includegraphics[width=4.8 in, height=3.4 in]{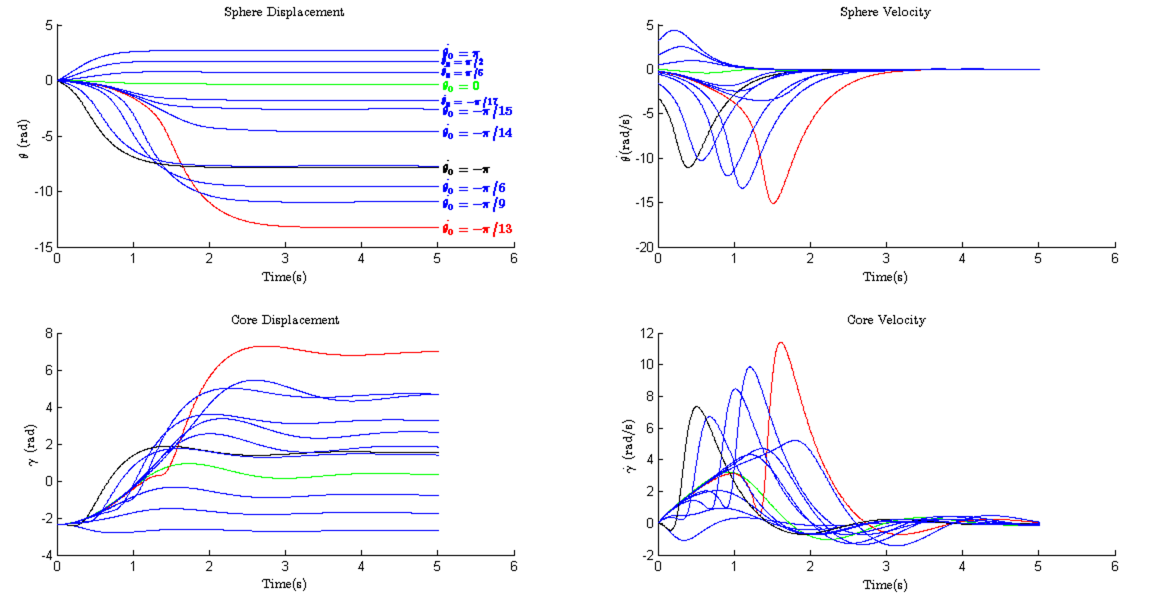}\newline(b)

\caption{(a) The case 1 motion and speed plots for $-\pi\leq\gamma_0<-\frac{\pi}{2}$, $X_0=[0\; \; 0\; \;  \gamma_0\; \;  0]$ values, (b) the case 2 motion and speed plots for $0\leq |\dot{\theta_0}| \leq \pi$ values, $X_0=[0\; \; \dot{\theta_0} \; \; -\frac{3\pi}{4}\; \;  0]$.}\label{Fig:Instabilitycorelocation}
\end{figure}
\subsubsection{Analytical Criteria}
We analyze the MM in instability region and effect of GB as section III.A through dynamics. The viscosity of core is $\zeta_\gamma= 0.01$ for first three cases (MM parts) and in Case 4 the viscous friction is increased to $\zeta_\gamma= 0.075$ to have clearer view over velocities due to defined dynamic constraint in inertia coefficient of GB path ($I_{c_{GB}}$).

\begin{figure}
\centering
\includegraphics[width=4.8 in, height=3.4 in]{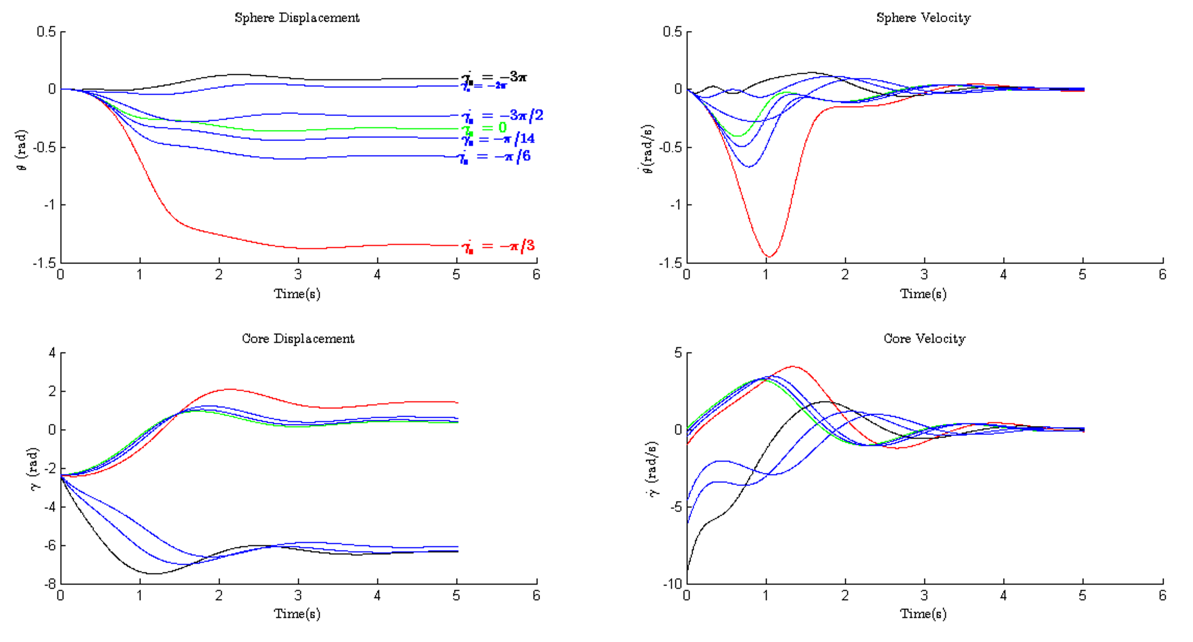}\newline(a)
\includegraphics[width=4.8 in, height=3.4 in]{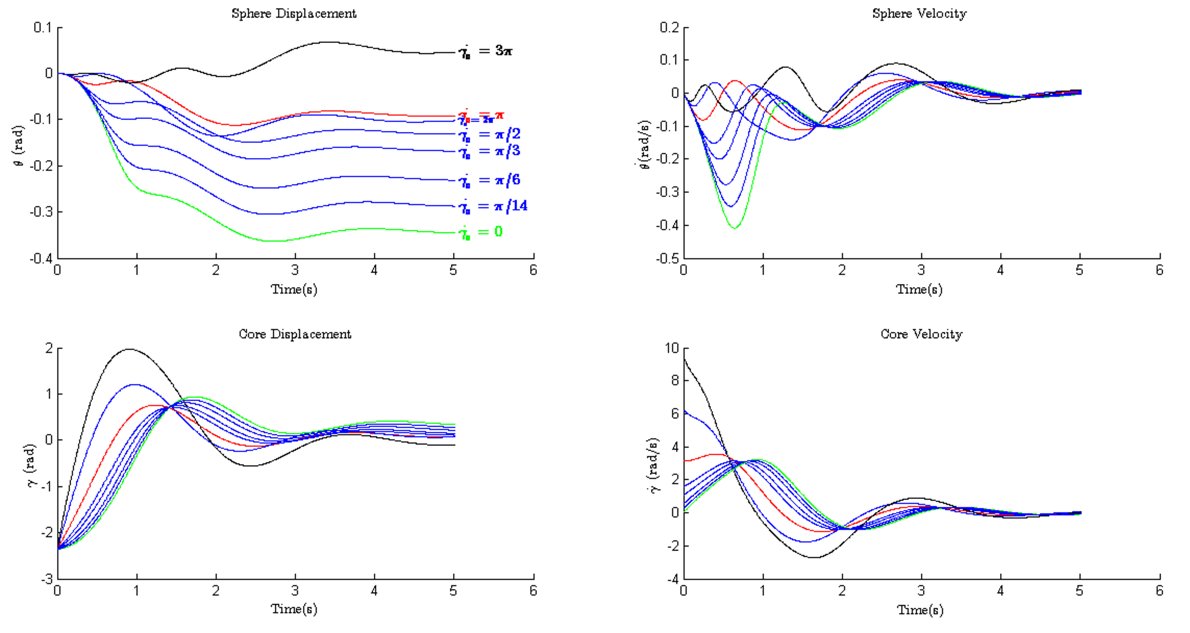}\newline(b)
\caption{The case 3 movement and velocity plots for $-3\pi\leq\dot{\gamma}_0<3\pi$ values, $X_0=[0\; \; 0\; \; -\frac{3\pi}{4} \; \; \dot{\gamma}_0]$. Note that the black, red and green lines stand for maximum initial input core velocity in both sides, unexpected nonlinear manner and no input velocity as reference , respectively. }\label{Fig:InstibilityCorespeedcontrol}
\end{figure}

\textbf{Case 1:}
The plot series in Fig.  \ref{Fig:Instabilitycorelocation}-a demonstrates the results for 6 different values of $\gamma_0$ in region of instability ($-\pi<\gamma_0\leq-\frac{\pi}{2}$). Mobility starts with basic initial condition $X_0=[\theta \; \; \dot{\theta}\;  \; \gamma \; \;\dot{\gamma}]=[0\; \; 0\; \;  \gamma_0\; \;  0]$. The red value ($\gamma_0=-\frac{\pi}{2}\;$ rad) is the critical point of motion since it aims the sphere to move in forward direction and also this is the place in which creates similar movement in lower half of sphere ($-\frac{\pi}{2}\leq\gamma_0\leq0$). We saw as initial location of the core climbs up, the total movement of sphere became backward. The $\gamma_0=-\frac{5\pi}{6}\;$ rad (Fig. \ref{Fig:Instabilitycorelocation} black line) is illustrating, displacement ratio jumps when the value gets nearer to the upper equilibrium point $\gamma_0=-\pi\;$ rad in stationary condition.

\textbf{Case 2:}
The system is excited with different values of sphere's initial velocities,$0\leq |\dot{\theta_0}| \leq \pi$. The core begins from fixed point in unstable region, $\gamma_0=-\frac{3\pi}{4}\;$ rad ($X_0=[0\;\dot{\theta_0} \; -\frac{3\pi}{4} \;  0]$). RollRoller robot follows direction of motion relative to sphere's speed. However, The range of locomotion or efficiency of movement interestingly is integrated. There is kinetic clue in movement, robot's core location over time depends to sphere's speed because the core is moving in circular path. As a result, it can be said the maximum displacement can be approximated as Eq. (\ref{Eq:speedCase2maximumdis}) in this case. This relation is not unit base relation.
\begin{equation}
\dot{\theta} \simeq \frac{\gamma}{g}
\label{Eq:speedCase2maximumdis}
\end{equation}

\textbf{Case 3:} Illustrating how SMR is able to maneuver despite the condition of actuator is one of the essential points. Fig. \ref{Fig:InstibilityCorespeedcontrol} shows locomotion of sphere with different velocity responses in both directions. The initial states were $X_0=[0\; \; 0 \; \; -\frac{3\pi}{4}\; \; 0\leq \dot{\gamma}_0 \leq 3\pi]$. In the state that robot contains counterclockwise velocity for certain level ($0\leq\dot{\gamma} \lessapprox -2\pi$) it produces backward locomotion. When the velocity exceeds this property, the $-2\pi \; $ rad/s, sphere's locomotion follows the opposite direction of same axis. However, specific constraint is been created for cases that core velocity is positive because the core location is at $\gamma_0 = -\frac{3\pi}{4}\;$ rad. Although, we inject positive velocity to core and expect clockwise turning for sphere (see Fig. \ref{Fig:InstibilityCorespeedcontrol}-b), the outcomes are showing this only possible when core's velocity exceeds the $2\pi\;$ rad/s. And, around the values over the $\pi$ rad/s it saturates around the red line ($\dot{\gamma}_0=\pi\;$ rad/s) in Fig. \ref{Fig:InstibilityCorespeedcontrol}. 

\textbf{Case 4:}
\begin{figure}
\centering
\includegraphics[width=4.8 in]{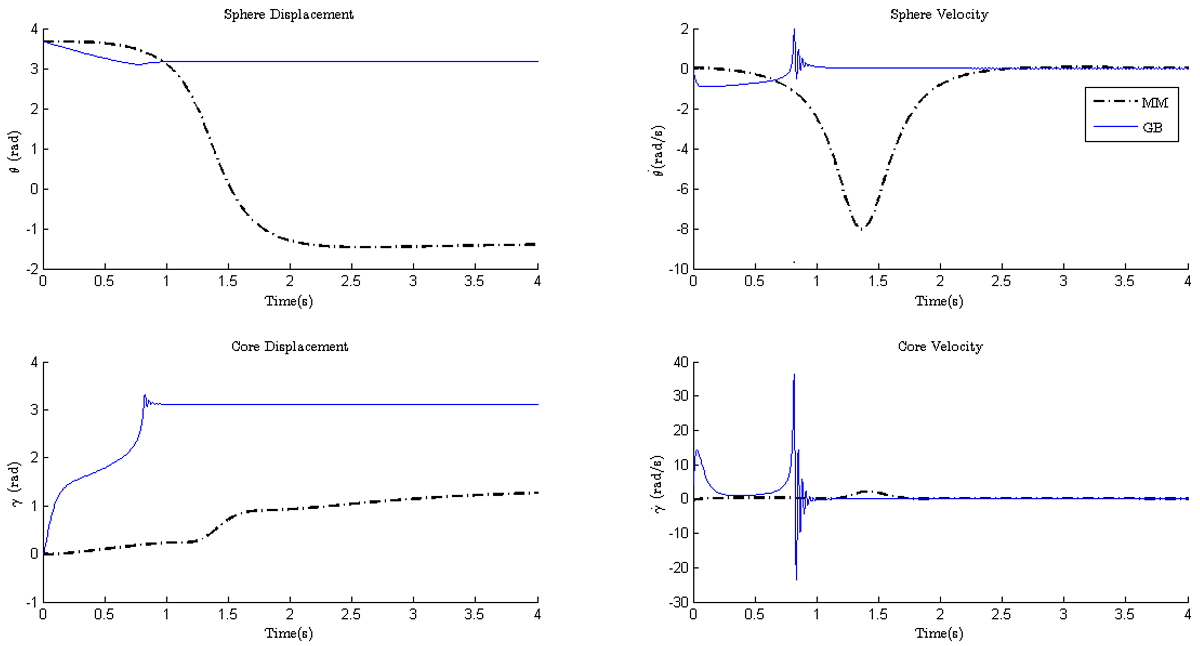}
\caption{The GB and MM behavior with initial condition as $X_0=[-\pi-\frac{\pi}{6}\; \; 0.1 \; \; 0 \; \;  -\frac{\pi}{8}]$}\label{Fig:GbTOPToBot}
\end{figure}
As it was obvious from Fig. \ref{Fig:InstibilityCorespeedcontrol}, in unstable regions, the problem of limiting the velocity of core (requiring sudden change of velocity parallel to clear negative impacts) to keep motion completely concentrated on one direction is become hard task. Fig. \ref{Fig:GbTOPToBot} gives an example point from case 4 and shows how GB outperforms itself and doesn't create certain backward locomotion. The most realistic results with containing impacts will be explained and shown in Section V. The Offset between the MM used model and GB is:
\begin{equation}
Offset=|\theta_{MM_{setteled}}|-|\theta_{GB_{setteled}}|=4.82 \; rad
\end{equation}
Also, the negative velocity that sphere carries, is totally omitted in GB. In this motion, core with fix circular turn in MM can create fluctuation in sphere displacement and also its velocity. Furthermore, the created ripple in core velocity at around $t=0.8 \;$sec is coming from estimation of inertia as we explained before. lastly, our algorithm in Section III.B can be seen as satisfying proposal for forward direction cycle with GB inclusion during high proportional core's speed (Fig. \ref{Fig:forwarddirectionmotion} cycles b,d and g).

\subsubsection{Forward Locomotion}
\begin{figure}[h!]
\centering
\includegraphics[width=4.8 in]{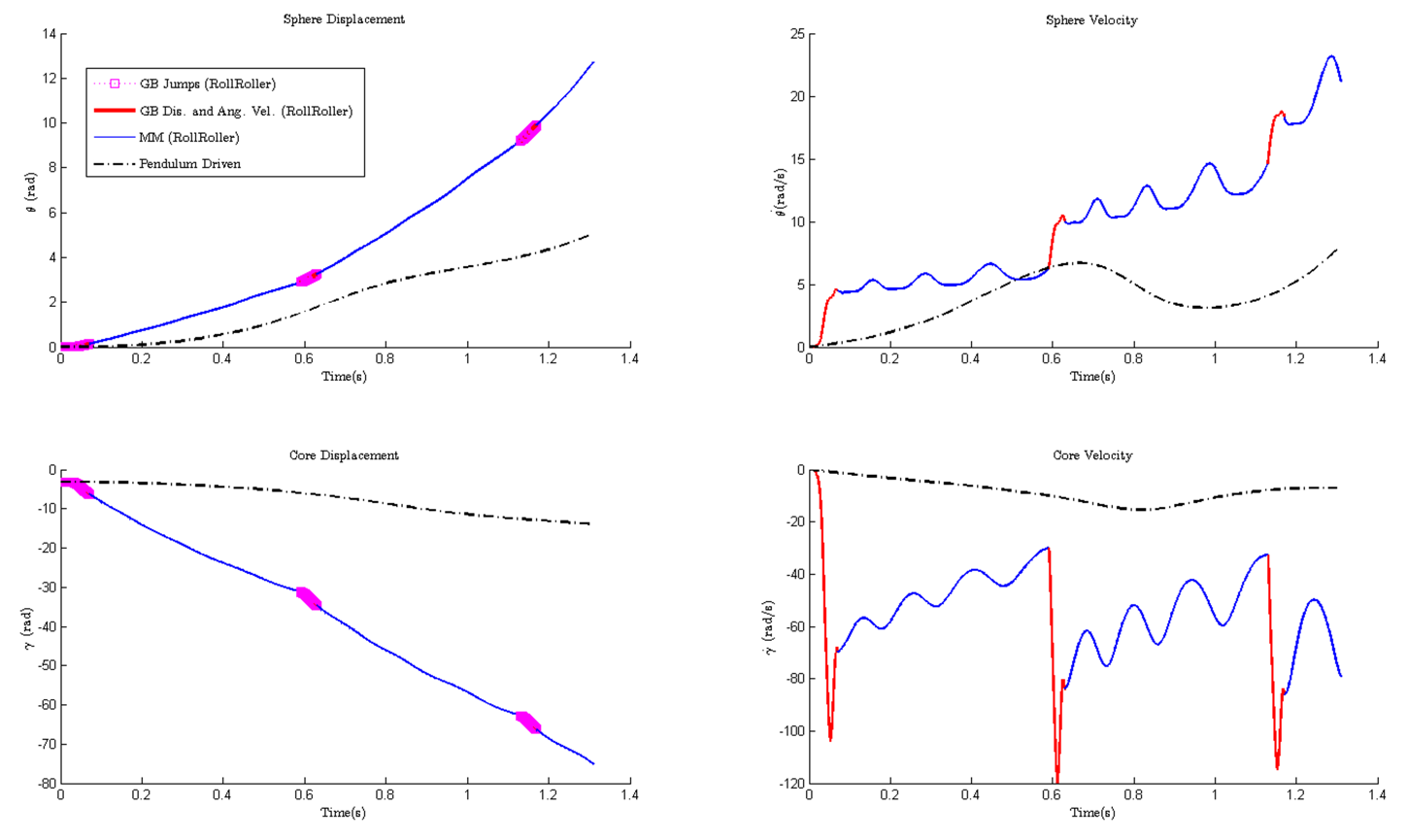}
\caption{The pendulum model which designed with our parameters and RollRoller maneuver in forward direction ,with $X_0=[0\; \; 0 \; \; -\frac{\pi}{2} \; \; 0]$ initial conditions.}
\label{Fig:ForwardDirectionAlgorithm}
\end{figure}

To see the compact manner of RollRoller by using the derived nonlinear Eqs. (\ref{Eq:motionmechanismsystem})-(\ref{Eq:Inertia}), the system has been taken to hybrid state. From the Section III.B, we consider the motion has all state properties from cycle a to h in Fig. \ref{Fig:forwarddirectionmotion} except cycle e. Cycle e only occurs in cases where core velocity is changing instantly. For hybrid model of GB and MM motion, we stem position control method from Algorithm \ref{Algo:forwarddirection} particularly in $\dot{\theta}<\dot{\gamma}$ condition. In our experiment $\eta_{\theta}$ and  $\eta_{\gamma}$ are approximately $\frac{\pi}{12}\;$ rad. However, We orientate the algorithm for implementation with over-writing its dependency on the sinusoidal locational form to have more freedom in our parameters variety. The constraints for line 7 are mutated as follows: 
\begin{multline}
\{ \cos(\theta) \leq -0.95 \; \; and \; \; \sin(\theta) \geq -0.2 \; \; and \; \; \dot{\theta} \geq 0 \; \; and \; \; \\
\dot{\gamma} \leq 0 
\; \; and \; \;  \cos(\gamma) \geq  0.9 \; \; \sin(\gamma) \leq 0.2 \; \; \}  \; \; OR 
\\
\{\cos(\theta) \geq 0.95 
\; \; and \; \; \sin(\theta)\leq 0.2 \; \; and \; \; \dot{\theta} \geq 0 \; \; and \; \; 
\\
\dot{\gamma}\leq 0 
\; \; and \; \; \cos(\gamma) \leq  -0.9  \; \; and  \; \; \cos(\gamma)\geq -0.2\} \\
\\ 
\label{ConstraintLabel1}
\end{multline}
Also, Line 11 is transformed to:
\begin{multline}
\{|\cos(|\gamma|)|\leq0.9 \; \; and \; \; \sin(|\gamma|)\leq 0\} \; OR
\\
\{|\cos(|\gamma|)|\leq0.9
\; \; and \; \; \sin(|\gamma|)\geq 0 \}
\label{ConstraintLabel2}
\end{multline}
Eqs. (\ref{ConstraintLabel1})-(\ref{ConstraintLabel2}) are designed for both $\alpha$ and $\beta$ regions. Also, sphere's small differential degree in clockwise direction in Algorithm 1 as shown in mutated equations is constrained by our design property that remove negative impacts in RollRoller motions when GB interacts (see Section V results). The hybrid integration of MM and GB pipes are defined by creating two series of core-sphere distances and viscous frictions. The viscous frictions of MM are $\zeta_{\theta_{MM}}=0.2$ and 
$ \zeta_{\gamma{MM}}=0.01$. Also, $\zeta_{\theta_{GB}}=0.2 $ and $ \zeta_{\gamma{GB}}=0.01$ are the GBs' viscous frictions. Fig. \ref{Fig:ForwardDirectionAlgorithm} shows our implemented algorithm for this motion with $X_0=[0 \; \; 0 \; \; -\frac{\pi}{2} \; \; 0]$ initial conditions. The step size of ODE45 solver is chosen $T=0.01$. The input torque $\tau_{\gamma}$ to core is $-0.075$ N$\cdot$m in counterclockwise direction in our experiment to satisfy Eq. (\ref{ComprassionLAandCore}).

Initially, RollRoller passes the core through GB pipe and the core in $\gamma_{jump_1} = -6.2304 $ rad leaves GB with $\dot{\gamma}_{jump_1} = -69.8284$ rad/s velocity. On the other side, pendulum-driven models (e.g., \cite{HalmeMotion1996,BicchiSpherenonholonomy,AugustJavadi2002,FirstPendulumBYQ2008,MITSchroll2008,VolvotSMR2011,DoublePendulumMahboubi2012,Controlfuzzy2013}) with given characteristics behave the locomotion by only circling the core with concave curve in MM. Our robot's velocity jump leads the motion from MM with semi-linear displacement. In the next step, the core entered GB for a second time ($\gamma_{jump_2}=-31.7397\;$ rad and $\theta_{jump_2}=2.9681\;$rad ), After exceeding $\dot{\theta}<\dot{\gamma}$. The second jump carries the $X=[3.2648 \; \; 9.9830 \; \; -34.7005 \; \;-83.9019]$ states to MM pipe. This iteration continues for next steps. However, This is position control for certain locomotion. For higher speeds, there must be integrated velocity control over the core. Simultaneously, robot to maintain/increase its speed, it may keep the core in GB for certain time. This limitation can come to sight after certain iterations.

The advantages of our model are visible from mathematical model results. The overall displacement without complete feedback control can create exponential increase in comparison to its own counterparts. 
\begin{equation}
Offset=|\theta_{RollRoller}|-|\theta_{Pendulum Driven}|=7.713 \; rad
\end{equation}
Also, the displacement is free from any swing during the time. We see the velocity response of our robot is rectified in negative part of sinusoidal. There is minor velocity diversity during the motion. The peak in core velocity during jump comes from low dissipation energy which can be minimized with feedback control within GB path. We didn't have a look to the motion of sphere in low velocities $\dot{\theta}\geq\dot{\gamma}$ since this motion is similar to its counterparts that actuates with depending to gravitational and torque-reaction forces.
\section{Numerical Simulation}
\begin{figure}
\centering
\includegraphics[width=3.5 in]{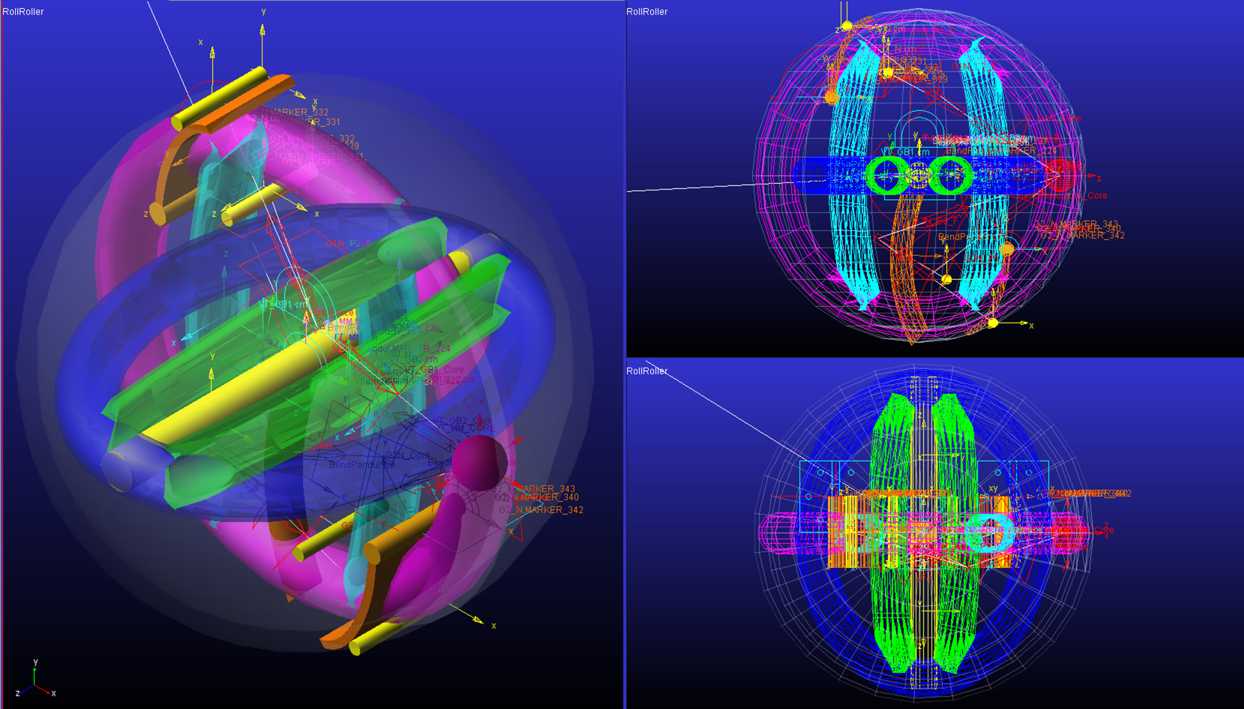}
\caption{Adams/View RollRoller model.}\label{Fig:AdamschematicRollRoller}
\end{figure}

To validate the derived nonlinear equations and get more insight to motion of RollRoller, the model is implemented to Adams/View simulation space in Fig. \ref{Fig:AdamschematicRollRoller}. We basically designed and imported the Solidworks model of RollRoller by utilizing the informations from Table \ref{Tab:NumericalVariabledyn}.

\begin{figure}
\centering
\includegraphics[width=1.7 in, height=.9 in]{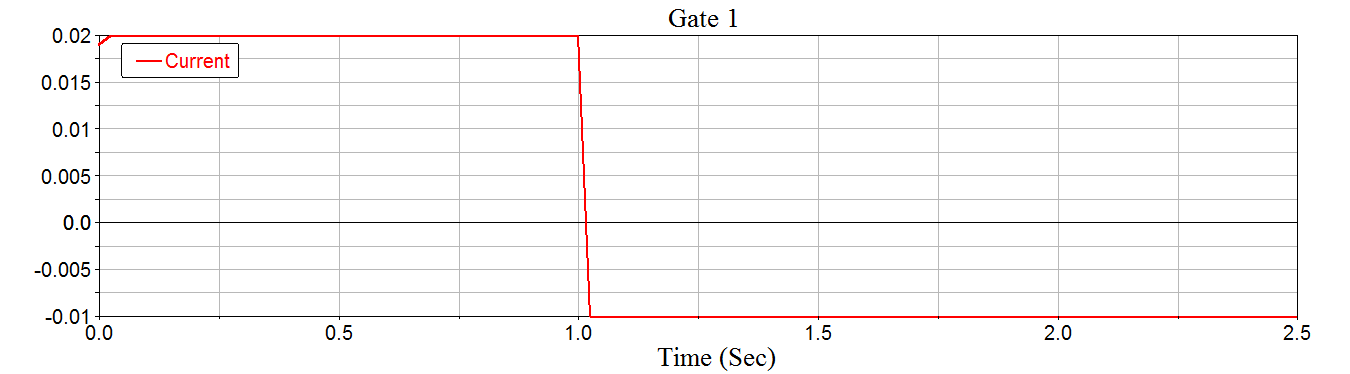}
\includegraphics[width=1.7 in, height=.9 in]{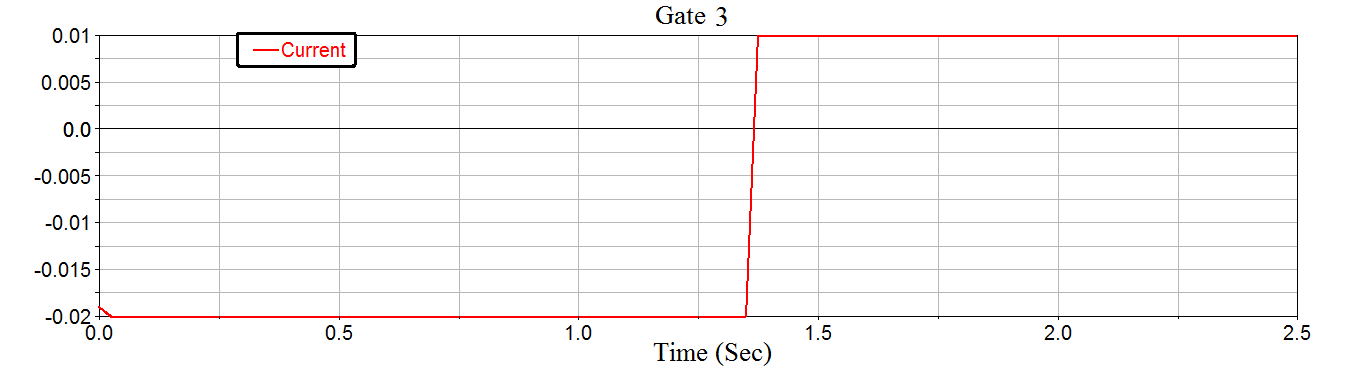}
\caption{Force verses time for gates functioning period.}\label{Fig:AdamGates}
\end{figure}

\begin{figure}
\centering
\includegraphics[width=1.16 in, height=.68 in]{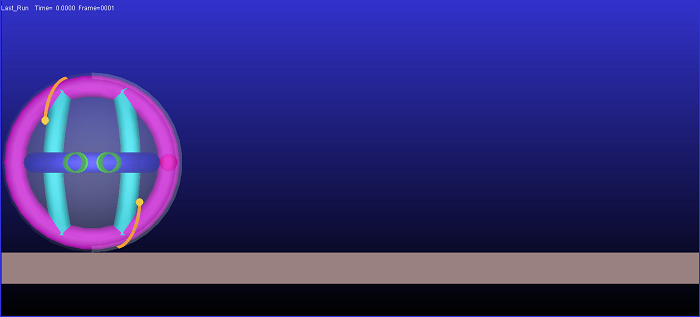} 
\includegraphics[width=1.16 in, height=.68 in]{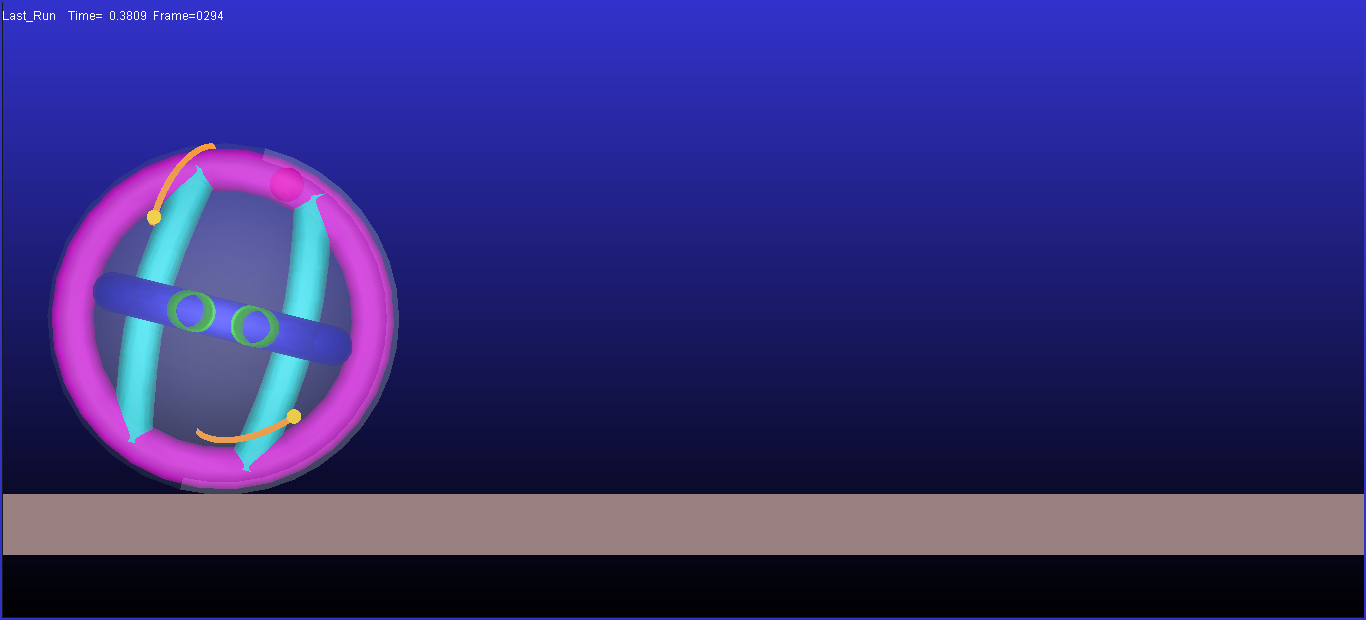} 
\includegraphics[width=1.16 in, height=.68 in]{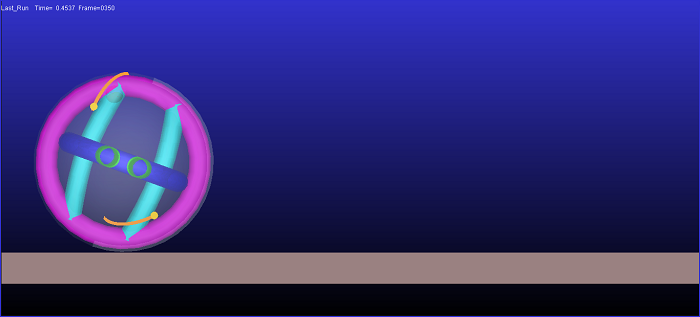} 
\includegraphics[width=1.16 in, height=.68 in]{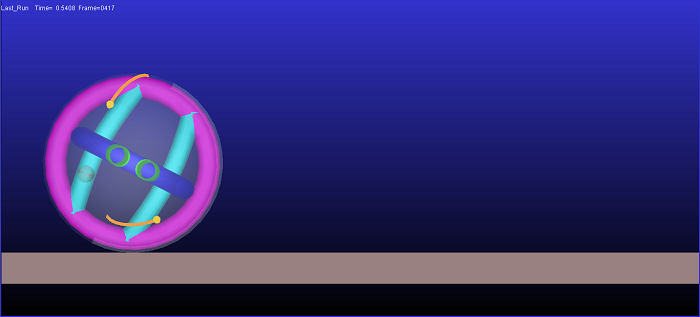} \linebreak 
 t = 0 sec 
\space\space\space\space\space\space\space\space\space\space\space\space\space t = 0.38 sec
\space\space\space\space\space\space\space\space\space\space\space t = 0.45 sec
\space\space\space\space\space\space\space\space\space\space\space t =  0.54 sec

\includegraphics[width=1.16 in, height=.68 in]{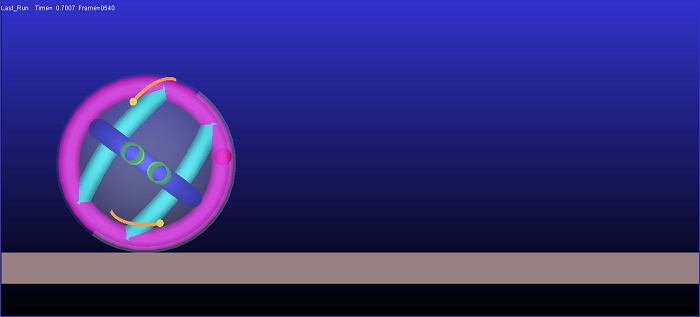}
\includegraphics[width=1.16 in, height=.68 in]{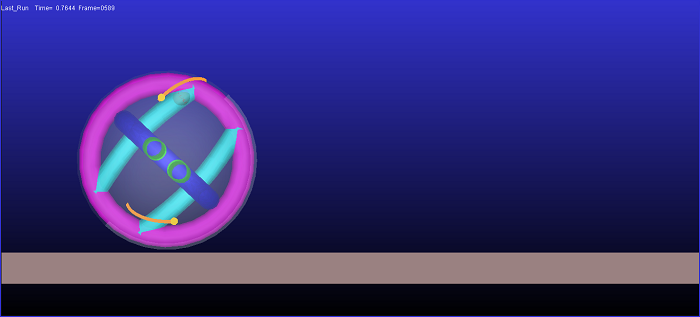}
\includegraphics[width=1.16 in, height=.68 in]{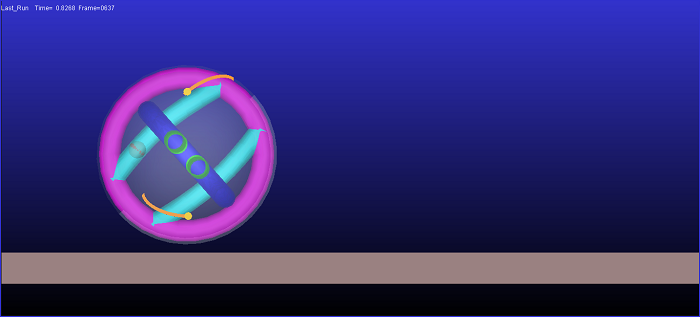}
\includegraphics[width=1.16 in, height=.68 in]{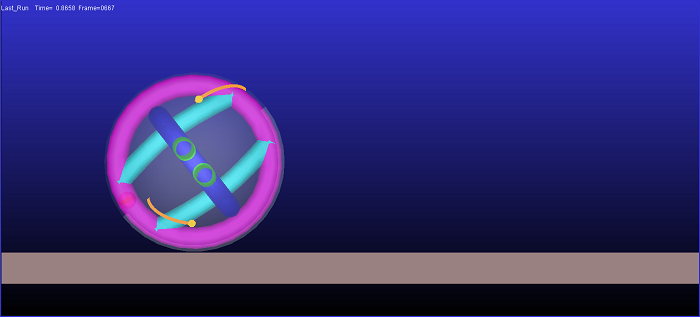} \linebreak
t = 0.7 sec 
\space\space\space\space\space\space\space\space\space\space\space\space\space t = 0.76 sec
\space\space\space\space\space\space\space\space\space\space\space t = 0.82 sec
\space\space\space\space\space\space\space\space\space\space\space t =  0.86 sec

\includegraphics[width=1.16 in, height=.68 in]{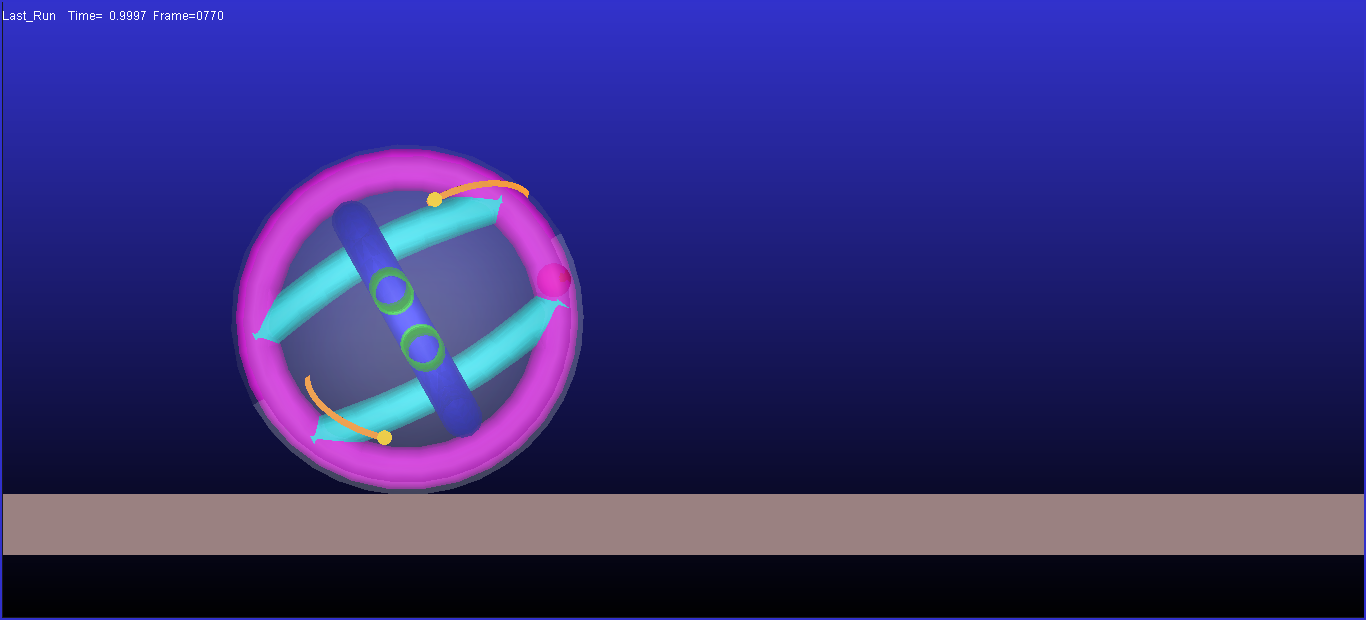} 
\includegraphics[width=1.16 in, height=.68 in]{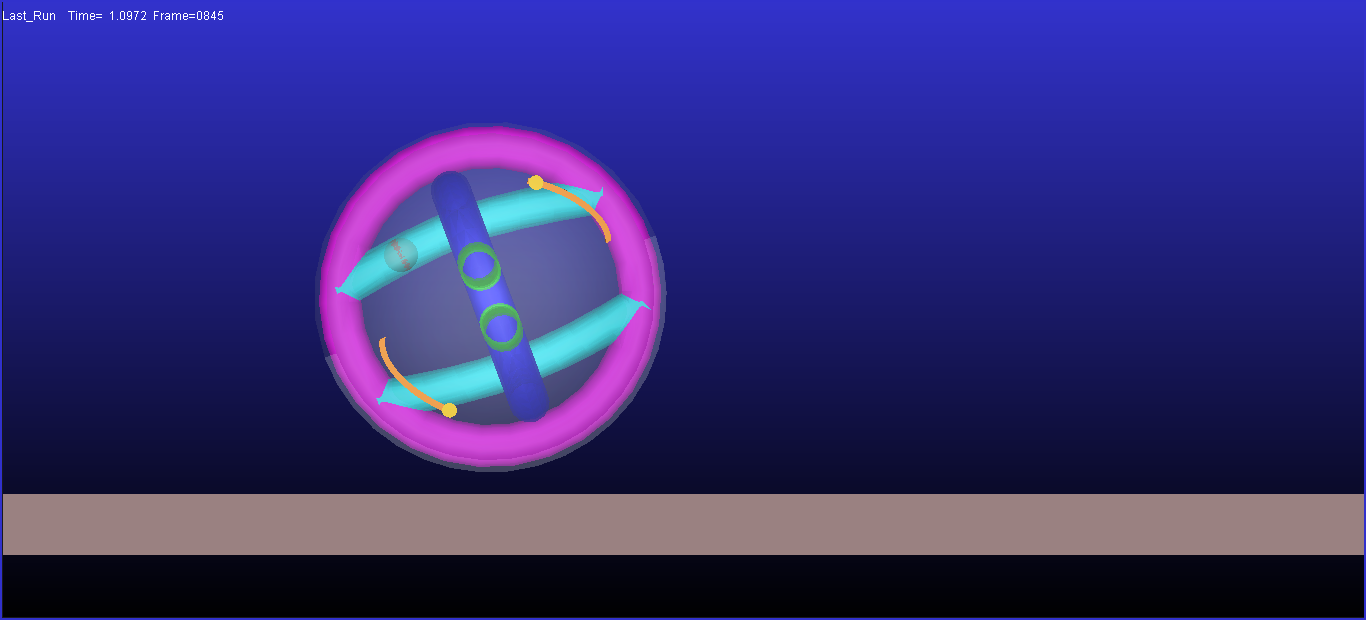}
\includegraphics[width=1.16 in, height=.68 in]{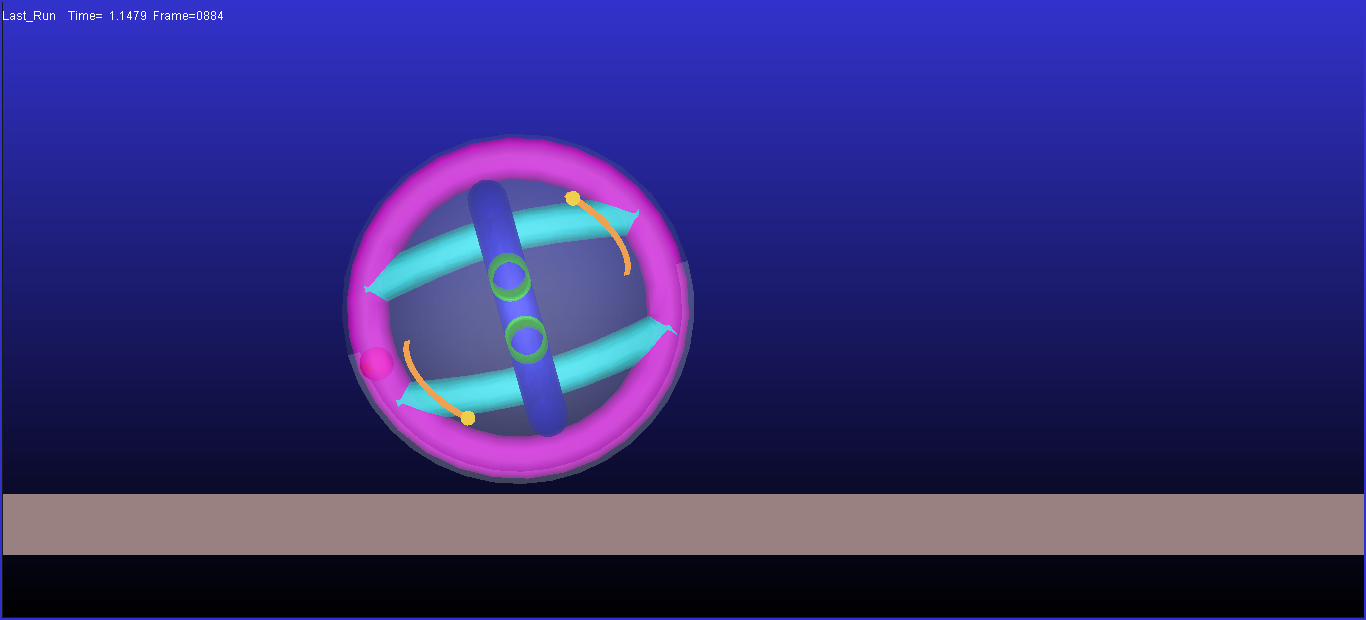}
\includegraphics[width=1.16 in, height=.68 in]{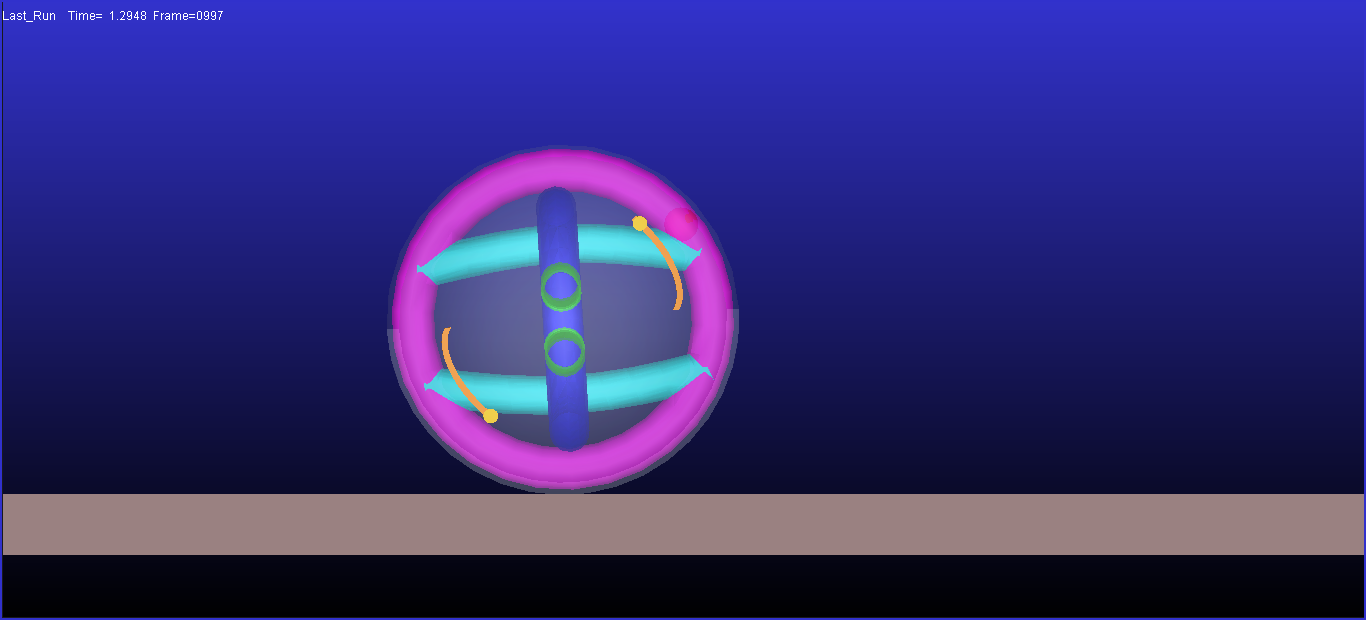}\linebreak
t = 0.99 sec 
\space\space\space\space\space\space\space\space\space\space\space\space\space t = 1.09 sec
\space\space\space\space\space\space\space\space\space\space\space t = 1.14 sec
\space\space\space\space\space\space\space\space\space\space\space t = 1.29 sec

\includegraphics[width=1.16 in, height=.68 in]{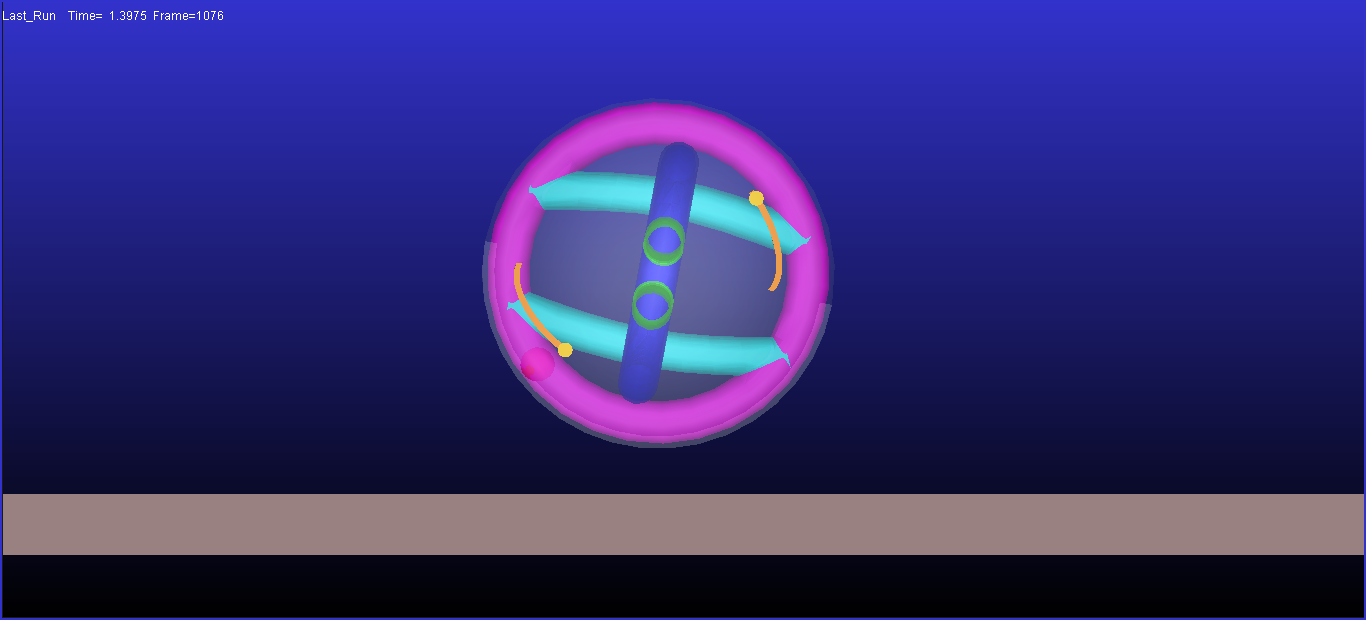}
\includegraphics[width=1.16 in, height=.68 in]{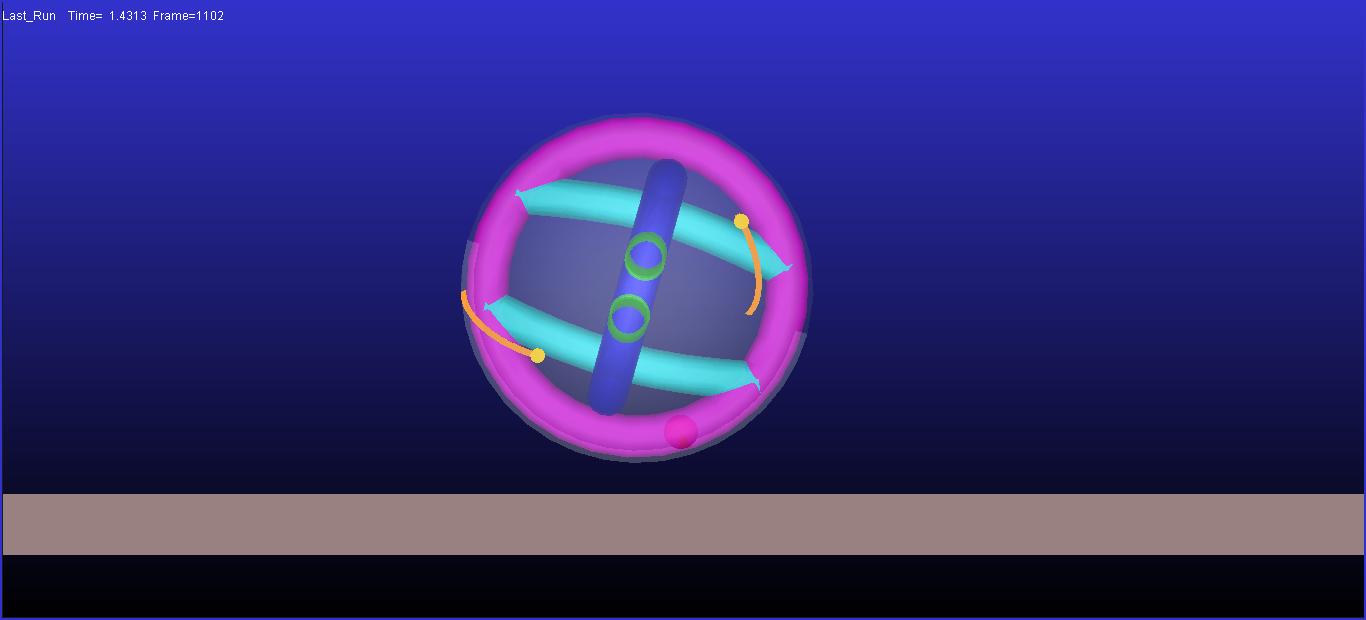}
\includegraphics[width=1.16 in, height=.68 in]{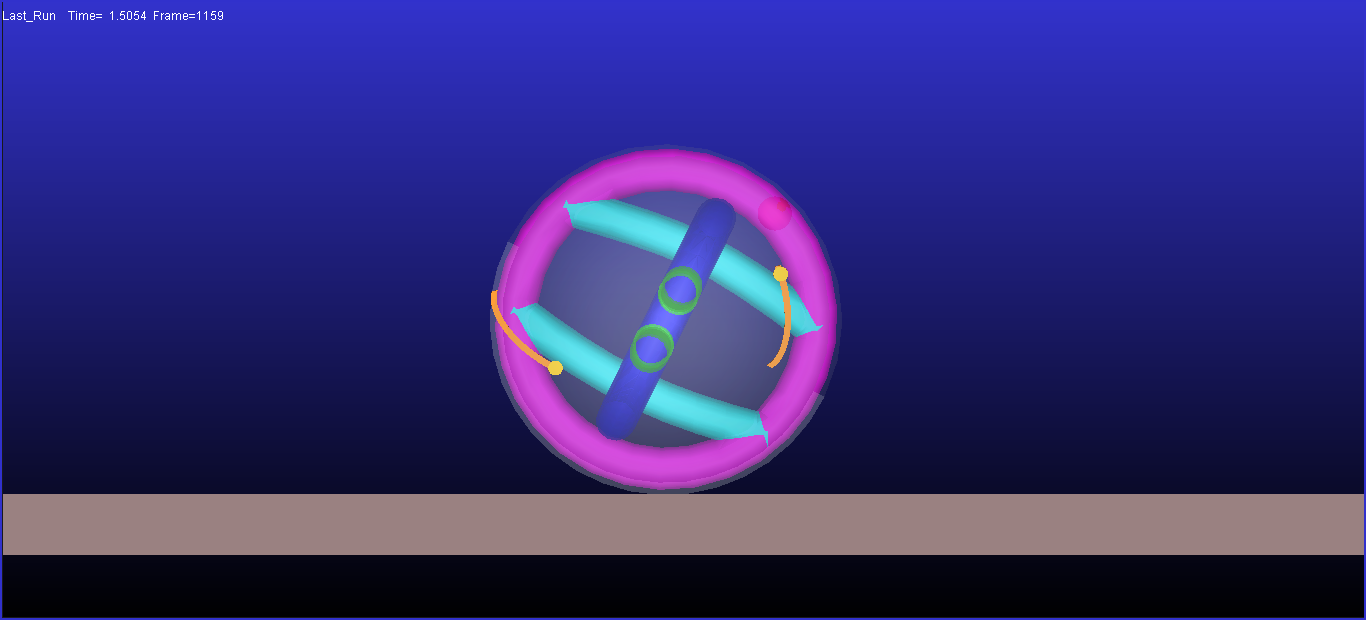}
\includegraphics[width=1.16 in, height=.68 in]{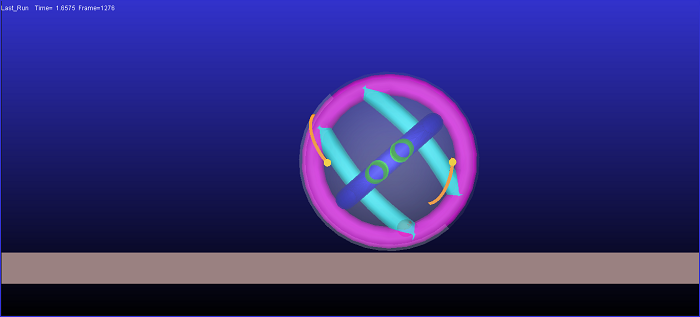}\linebreak
t = 1.39 sec 
\space\space\space\space\space\space\space\space\space\space\space\space\space t = 1.43 sec 
\space\space\space\space\space\space\space\space\space\space\space t = 1.5 sec
\space\space\space\space\space\space\space\space\space\space\space t = 1.65 sec 

\includegraphics[width=1.16 in, height=.68 in]{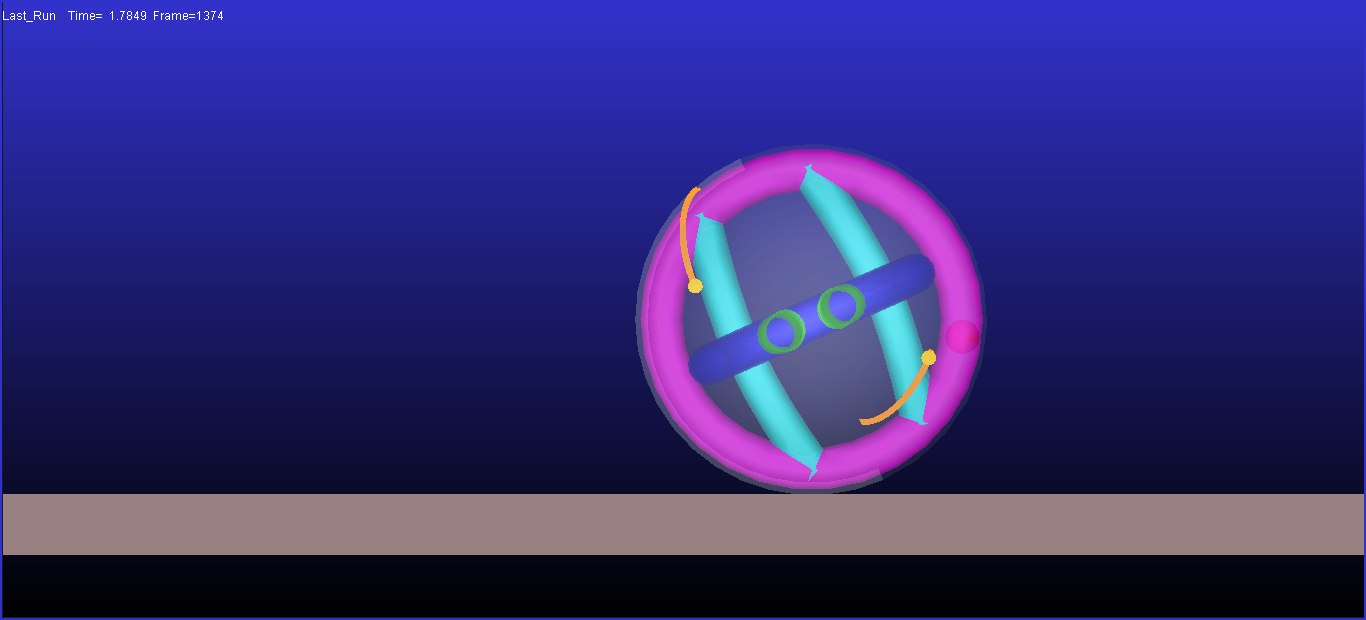}
\includegraphics[width=1.16 in, height=.68 in]{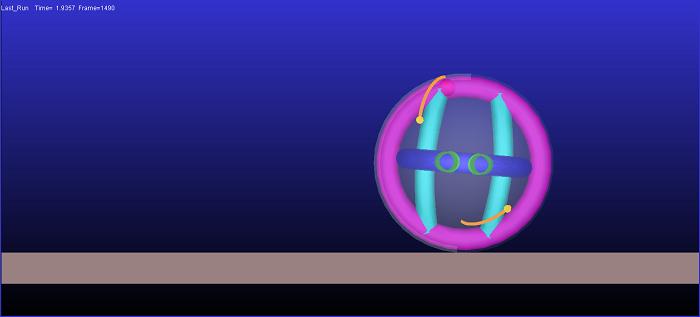}
\includegraphics[width=1.16 in, height=.68 in]{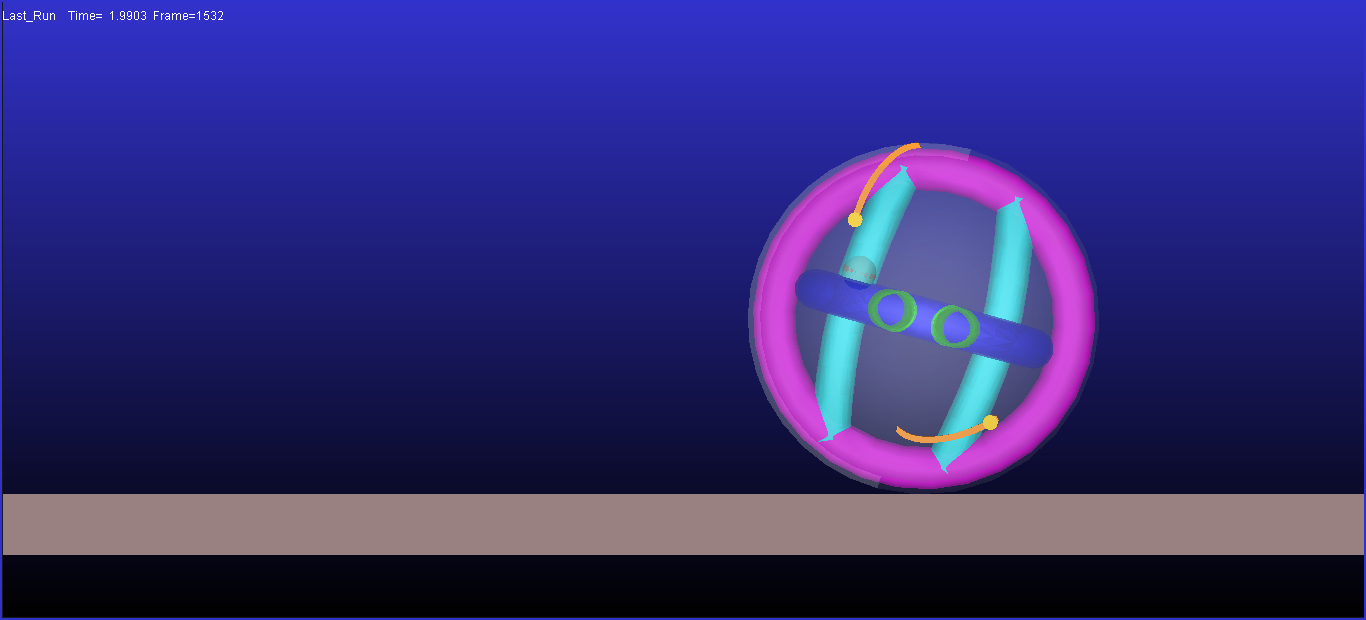}
\includegraphics[width=1.16 in, height=.68 in]{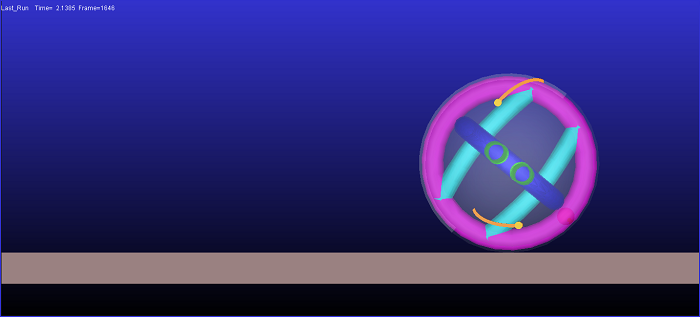}\linebreak
t = 1.78 sec 
\space\space\space\space\space\space\space\space\space\space\space\space\space t = 1.93 sec 
\space\space\space\space\space\space\space\space\space\space\space t = 1.99 sec
\space\space\space\space\space\space\space\space\space\space\space t = 2.1 sec

\includegraphics[width=1.16 in, height=.68 in]{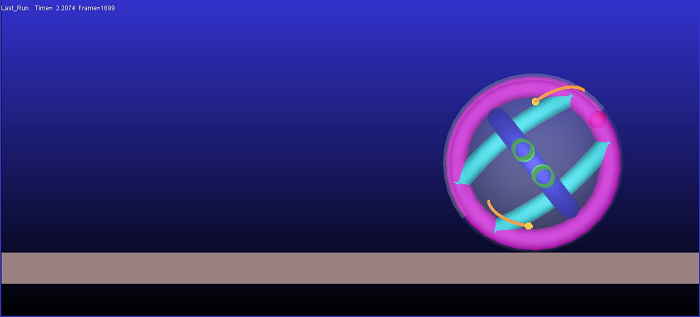}
\includegraphics[width=1.16 in, height=.68 in]{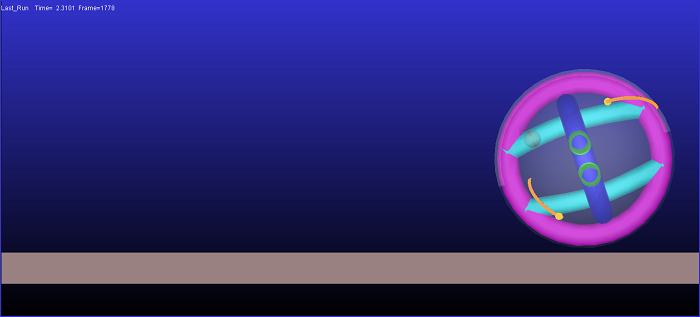}
\includegraphics[width=1.16 in, height=.68 in]{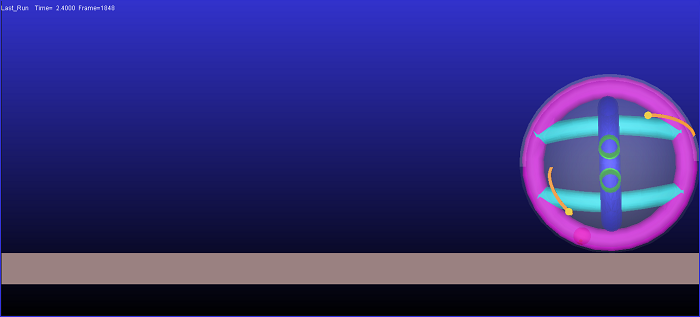}
\linebreak
t = 2.2 sec 
\space\space\space\space\space\space\space 
\space\space\space\space\space\space t = 2.31 sec
\space\space\space\space\space\space\space\space\space\space\space\space t = 2.4 sec
\caption{Captured frames in forward locomotion in Adams/view simulation within $2.5\;\;$sec.}\label{Fig:Adamsviewshots}
\end{figure}
Our created model is estimated the superfluid circulation to the pendulum-like pusher to move the core. The friction property of sphere-ground also is adjusted from dissipation function in mathematical model as $0.6$ for static friction and $0.3$ for dynamic friction. This friction coefficient indicates the existence of certain slippage between the sphere and plain surfaces which makes the model more realistic in application. The mass of spherical shell is considered about $900\;\;grams$ and rest of weight is spread around the parts to have homologous mass distribution over system. We assume the core and pipe surfaces are in minimum friction (approx. $0$). Two of the control gates are designed with curvy model to minimize the reaction force when core pass this object. In this simulation, control gates are inspired from Fig. \ref{Fig:mechanicandgate} (b) with difference that they are articulating instead of sliding (two sliding gates are replaced with one articulating gate) for simplicity. Lastly, the FT core's mass is designed by means of Eq. (\ref{masscoreandsphererelation}) as same as dynamic model ($0.25\;\;$kg). 
\subsection{Exceeded Core Velocity ($\dot{\theta}<\dot{\gamma}$)}
To see and compare our model with and without the algorithm, we design simulated RollRoller to move in compact form in $\dot{\theta}<\dot{\gamma}$ condition.  We excite the pusher with $0.35$ N which creates clockwise torque to push the core. 
The functioning period for both gates 1 and 3 are like in Fig. \ref{Fig:AdamGates}. These step inputs help us to achieve the relative response in examining the Algorithm \ref{Algo:forwarddirection}. Our SMR begins the locomotion with $X_0=[0 \; \; 0 \; \; -\frac{\pi}{2} \; \; 0]$ initial conditions. In Fig. \ref{Fig:Adamsviewshots}, we have captured some frames after running the Adams/View simulation.

\begin{figure}
\centering
\includegraphics[width=4.8 in]{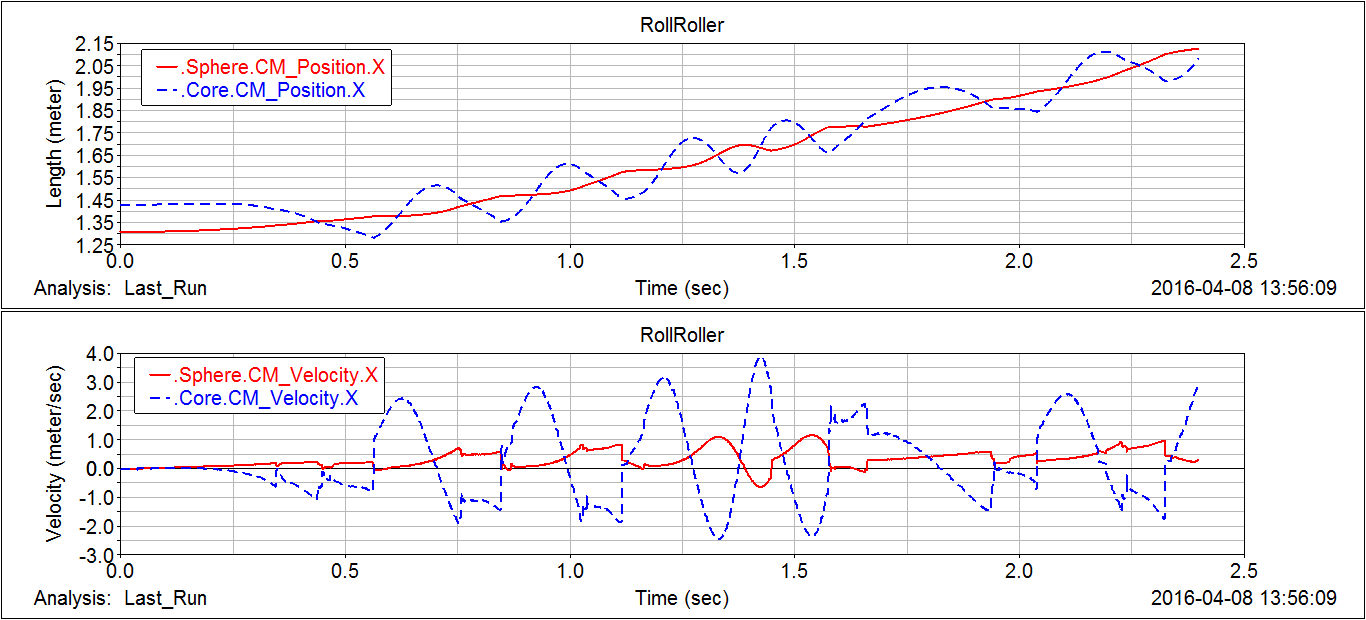}
\caption{Movement and velocity of core and sphere for 2.5 sec.}\label{Fig:Adams/Viewdispandvel}
\end{figure}

It is observed, RollRoller moves in linear form during the transient state before t = $0.5\;$sec (see Fig. \ref{Fig:Adams/Viewdispandvel}). 
In around $0.45-0.54 \;$sec intervals, core passes the GB to prevent the negative velocity and backward locomotion of sphere. However, the rectified velocity can be seen more clearly about $0.75 \;$ sec when GB is used for a second time. After letting the core to circulate by algorithm, we change the function of gates to open just MM pipes.
Between $1.09$ to $1.5 \;$sec, the fluctuation is appeared in core displacement and velocity, consequently. In the next stage, we set back the algorithm to observe the morphology in different state conditions after $1.5\;\;$sec. Therefore, we first let core enter GB in the beginning with $0<\theta_e<k\pi+\frac{\pi}{3}$ various conditions and then we change it to $k\pi-\frac{\pi}{3}\leq \theta_e<0$. The results show us although we succeed in our proposal method in this example, the advantages and sensitivity of algorithm become more clear. Y axis jumps and decrease in conserved velocity of sphere are relative to the time that core passes from MM to GB through gates.

\begin{figure}[h!]
\centering
\includegraphics[width=3.5 in]{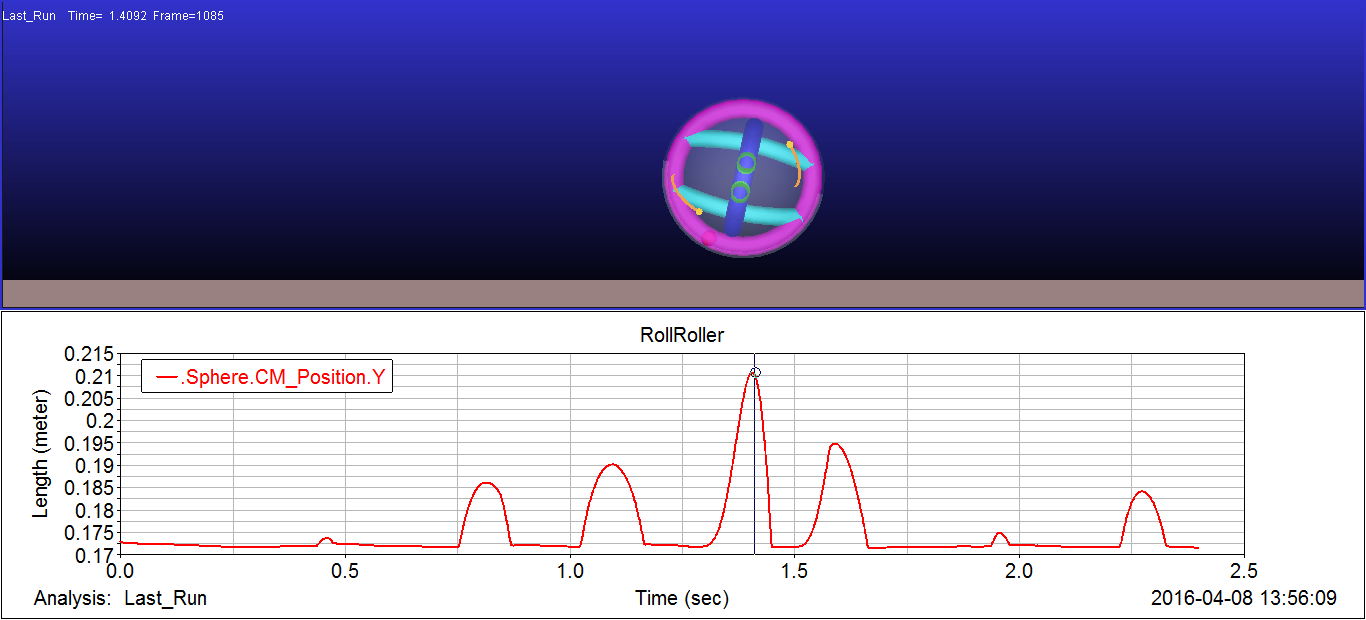}
\caption{RollRoller jumps in played simulation. Note: 
The $21\;\;$cm as the largest value before 1.5 sec stands for robots without the algorithm \ref{Algo:forwarddirection}.}\label{Fig:Adam/viewJumpadamYaxis}
\end{figure}

It is observed, as we proposed the algorithm to let robot passes the core from the GB when sphere is in $0<\theta_e<k\pi+\frac{\pi}{3}$, firstly it stops the sphere to have backward motion. Secondly, as we tend to have sharper degrees (i.e. $k\pi+\frac{\pi}{3}<\theta_e<k\pi+\frac{\pi}{2}$) in sphere and simultaneously requires the core to pass the GB in these locations, there are relative jumps in y axis in RollRoller (Fig. \ref {Fig:Adam/viewJumpadamYaxis}). Therefore, the defined constraints in Algorithm \ref{Algo:forwarddirection} matches the optimized movement criteria. Moreover, entering the core to GB in $k\pi-\frac{\pi}{3}\leq \theta_e<0$ locations creates smoother motion in x axis but it has negative impacts in velocity of robot in which decreases our positive locomotion (see Fig. \ref{Fig:Adams/Viewdispandvel}, after $1.6 \;\;sec$). The reason is the created reaction force that gates gives to total sphere when core pases GB. Next, the difference of mathematical model and simulated SMR in the acceleration become more understandable.In this simulation, We are not able to see major increase in velocity of RollRoller since the imperfect form of curvy gates cancel the great deal of energy by injecting the reaction force to pusher (pusher is connected to sphere's body via revolute joint). This problem would be nominal if the core was excited with fluid circulation by cylinders.

\begin{figure}[h!]
\centering
\includegraphics[width=3.5 in]{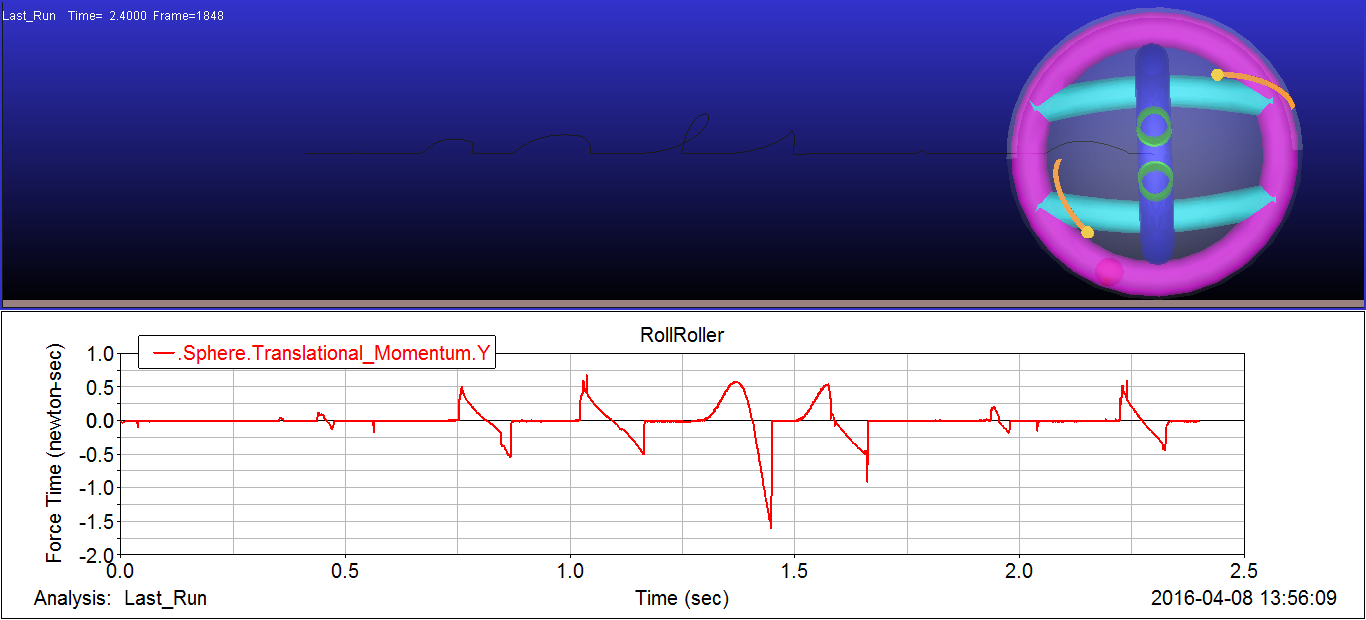}
\caption{RollRoller jumps in played simulation. Note: $3.8 \;$cm value before $1.5\;\;$sec highlights the largest jump during the deactivation of RollRoller's designed motion algorithm (Algorithm \ref{Algo:forwarddirection}).}\label{Fig:Adam/viewMomentumTrans}
\end{figure} 
We see from Fig. \ref{Fig:Adam/viewJumpadamYaxis}-\ref{Fig:Adam/viewMomentumTrans} that our translational momentum in y axis and range of jumps (Fig. \ref{Fig:Adam/viewJumpadamYaxis}: $t_1 \;=\; 0.45$ sec , $y_1 \; = \; 0.18\;$cm and $t_2 \;=\; 1.95$ sec , $y_2\;=\; 0.27$ cm) worked out perfectly inside the created algorithmic constraints. Moreover, The greatest deal of momentum energy diversity and highest jumps happens during time (t = 1.4 sec ) that robot's algorithm is deactivated.
\subsection{Turning Locomotion}
\begin{figure}[h!]
\centering
\includegraphics[width=3.8 in]{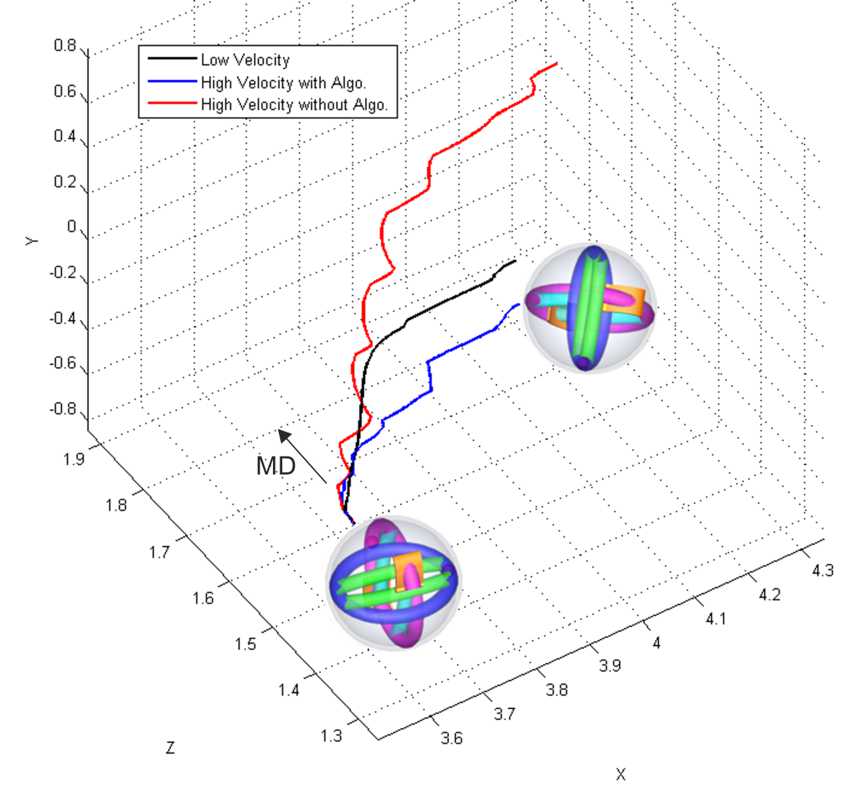}
\caption{Robot during its turning action for three methods. Note: The low velocity (black line), high velocity with algo. (Blue line) and high velocity without algo. (red line) stand for low core's speed ($\dot{\theta}\geq\dot{\gamma}$), high core's speed with algorithm ($\dot{\theta}<\dot{\gamma}$) and high core's speed without algorithm ($\dot{\theta}<\dot{\gamma}$). }\label{Fig:TurningMotionAdam}
\end{figure}
To see the robot when it's TT's section activated for 3D locomotion, we prepare the system for turning action as an example. By citing the Section III.C, the TT core's mass is $0.26 \;kg$ with same body properties as FT core. As an important structural characteristic before the simulation, the TT core is blocked in same place as Fig. \ref{Fig:CircularTurningSchematic} within $^{+}_{-}\;1\;$cm space. Also to prevent severe vibration in robot while it is turning, the damping and stiffness factors of contact between blockers and TT core are $10^{4} \;$N$\cdot$s/m and $10^{3}$ N/m, respectively.

We see the results in Fig.  \ref{Fig:TurningMotionAdam} for three separate scenarios. The simulation is run for $t=1.7 \;$sec with 0.0012 step size. When RollRoller wants to have turn in low velocity ($F_{c_{T}} > F_{c_{Max}}$), the behavior is smooth and lets it to have $90^o$ turn without any swing (pusher input force = $-0.24\;$N). However, by exceeding the core speed depending to sphere's speed (pusher input force = $-0.35 \;$N), we achieve  resealable movements.  Serious swings take place in its transient response while our algorithm is deactivated (i.e., pendulum driven models general problems) but motion with algorithm gives us clear explanation about ability to control the sphere in higher velocities. Not only robot can reach the required position in lesser time but also it does not contain fluctuation in its locomotion. The only problem as explained in Section V.A is the imperfect gates that wastes some energies when core carrying, becomes a deterrent force to sphere via pusher.

\section{Conclusion}
In this paper, we introduce mechanism of RollRoller as a novel SMR that works with high density fluid circulation. Instead of having mechanisms that solely rely on single or integration of previous models for their actuations, RollRoller robot includes all motivation methods in a single unique model with 4 degrees of freedom. We omit the common unstable region of pendulum-like SMR drivers with our mechanism and validate the results via derived nonlinear dynamics. By using the gained insight to the locomotion manners of RollRoller, we show how our proposed algorithmic motion control outperforms through nonlinear dynamics. Next, we import our model to Adams/View simulator and observe how backward jumpings are removed during high velocities beside the improvements in their fluctuations. Our results demonstrate that RollRoller like wheeled robots, is now able to accelerate in the field which solves the another main problem in ball shape robots. To illustrate the one of degrees of freedom and strengthen the proposed motion planning in 3D plain, we check the performance of RollRoller while it is turning. Despite some simulation restrictions that come from our actuators, we see that RollRoller is able to have about $90^o$ turn in most linear and the fastest manner. Although, these results raise an interesting point about spherical mobile robots, they show requirements in advancing simulation space for hydraulic application. Also, Our future primary aim is to create a general non-holonomic control formulation beside velocity stabilizer for the system.

%





\section*{Acknowledgment}
The authors would like to thank Dr. Larissa Khodadadi and Negar Esmaeilzadeh Kandjani for their contributions in preparing this paper. 




\section*{References}

\bibliography{references} 

\begin{thebibliography}{10}
\expandafter\ifx\csname url\endcsname\relax
  \def\url#1{\texttt{#1}}\fi
\expandafter\ifx\csname urlprefix\endcsname\relax\def\urlprefix{URL }\fi
\expandafter\ifx\csname href\endcsname\relax
  \def\href#1#2{#2} \def\path#1{#1}\fi

\bibitem{RollingNature}
R.~H. Armour, J.~F. Vincent, Rolling in nature and robotics: A review, Journal
  of Bionic Engineering 3~(4) (2006) 195 -- 208.
\newblock \href {http://dx.doi.org/10.1016/S1672-6529(07)60003-1}
  {\path{doi:10.1016/S1672-6529(07)60003-1}}.

\bibitem{PolarNASA}
A.~Behar, J.~Matthews, F.~Carsey, J.~Jones, Nasa/jpl tumbleweed polar rover 1
  (2004) 395.
\newblock \href {http://dx.doi.org/10.1109/AERO.2004.1367622}
  {\path{doi:10.1109/AERO.2004.1367622}}.

\bibitem{Tworigidrolling}
Z.~Li, J.~Canny, Motion of two rigid bodies with rolling constraint, IEEE
  Trans. Robot. Autom. 6~(1) (1990) 62--72.
\newblock \href {http://dx.doi.org/10.1109/70.88118}
  {\path{doi:10.1109/70.88118}}.

\bibitem{HalmeMotion1996}
A.~Halme, T.~Schonberg, Y.~Wang, Motion control of a spherical mobile robot 1
  (1996) 259--264.
\newblock \href {http://dx.doi.org/10.1109/AMC.1996.509415}
  {\path{doi:10.1109/AMC.1996.509415}}.

\bibitem{BicchiSpherenonholonomy}
A.~Bicchi, A.~Balluchi, D.~Prattichizzo, A.~Gorelli, Introducing the
  "sphericle": an experimental testbed for research and teaching in nonholonomy
  3 (1997) 2620--2625.
\newblock \href {http://dx.doi.org/10.1109/ROBOT.1997.619356}
  {\path{doi:10.1109/ROBOT.1997.619356}}.

\bibitem{BrownCMUGyrosignlewheel}
J.~Brown, H.B., Y.~Xu, A single-wheel, gyroscopically stabilized robot 4 (1996)
  3658--3663.
\newblock \href {http://dx.doi.org/10.1109/ROBOT.1996.509270}
  {\path{doi:10.1109/ROBOT.1996.509270}}.

\bibitem{Bhattacharyarolling2000}
S.~Bhattacharya, S.~Agrawal, Spherical rolling robot: a design and motion
  planning studies, IEEE Trans. Robot. Autom. 16~(6) (2000) 835--839.
\newblock \href {http://dx.doi.org/10.1109/70.897794}
  {\path{doi:10.1109/70.897794}}.

\bibitem{AugustJavadi2002}
A.~H.~J. A, P.~Mojabi, Introducing august: a novel strategy for an
  omnidirectional spherical rolling robot 4 (2002) 3527--3533.
\newblock \href {http://dx.doi.org/10.1109/ROBOT.2002.1014256}
  {\path{doi:10.1109/ROBOT.2002.1014256}}.

\bibitem{MITSchroll2008}
G.~C. Schroll, Design of a spherical vehicle with flywheel momentum storage for
  high torque capabilities, Master's thesis, Massachusetts Institute of
  Technology, Cambridge (2008).

\bibitem{FirstPendulumBYQ2008}
D.~Liu, H.~Sun, Q.~Jia, Stabilization and path following of a spherical robot
  (2008) 676--682\href {http://dx.doi.org/10.1109/RAMECH.2008.4681358}
  {\path{doi:10.1109/RAMECH.2008.4681358}}.

\bibitem{VolvotSMR2011}
M.~Ishikawa, R.~Kitayoshi, T.~Sugie, Volvot : A spherical mobile robot with
  eccentric twin rotors (2011) 1462--1467\href
  {http://dx.doi.org/10.1109/ROBIO.2011.6181496}
  {\path{doi:10.1109/ROBIO.2011.6181496}}.

\bibitem{Omnicron2012AIM}
C.~{Wei-Hsi}, C.~{Ching-Pei}, Y.~{Wei-Shun}, L.~{Chang-Hao}, L.~{Pei-Chun},
  Design and implementation of an omnidirectional spherical robot omnicron
  (2012) 719--724\href {http://dx.doi.org/10.1109/AIM.2012.6266036}
  {\path{doi:10.1109/AIM.2012.6266036}}.

\bibitem{GeometryPhase2005}
G.~Oriolo, M.~Vendittelli, A framework for the stabilization of general
  nonholonomic systems with an application to the plate-ball mechanism, IEEE
  Trans. Robot. 21~(2) (2005) 162--175.
\newblock \href {http://dx.doi.org/10.1109/TRO.2004.839231}
  {\path{doi:10.1109/TRO.2004.839231}}.

\bibitem{GeometryPhase2014}
A.~Morinaga, M.~Svinin, M.~Yamamoto, A motion planning strategy for a spherical
  rolling robot driven by two internal rotors, IEEE Trans. Robot. 30~(4) (2014)
  993--1002.
\newblock \href {http://dx.doi.org/10.1109/TRO.2014.2307112}
  {\path{doi:10.1109/TRO.2014.2307112}}.

\bibitem{GeometryPhase2015}
V.~Muralidharan, A.~D. Mahindrakar, Geometric controllability and stabilization
  of spherical robot dynamics, IEEE Trans. Automat. Control 60~(10) (2015)
  2762--2767.
\newblock \href {http://dx.doi.org/10.1109/TAC.2015.2404512}
  {\path{doi:10.1109/TAC.2015.2404512}}.

\bibitem{Algorithmbase2008}
M.~Svinin, S.~Hosoe, Motion planning algorithms for a rolling sphere with
  limited contact area, IEEE Trans. Robot. 24~(3) (2008) 612--625.
\newblock \href {http://dx.doi.org/10.1109/TRO.2008.921568}
  {\path{doi:10.1109/TRO.2008.921568}}.

\bibitem{Algorithmbase2010}
F.~Alouges, Y.~Chitour, R.~Long, A motion-planning algorithm for the
  rolling-body problem, IEEE Trans. Robot. 26~(5) (2010) 827--836.
\newblock \href {http://dx.doi.org/10.1109/TRO.2010.2053733}
  {\path{doi:10.1109/TRO.2010.2053733}}.

\bibitem{Controlnonholonomic2012}
A.~V. Borisov, A.~A. Kilin, I.~S. Mamaev,
  \href{http://dx.doi.org/10.1134/S1560354712030045}{How to control chaplygin's
  sphere using rotors}, Reg. and Chaotic Dynam. 17~(3) (2012) 258--272.
\newblock \href {http://dx.doi.org/10.1134/S1560354712030045}
  {\path{doi:10.1134/S1560354712030045}}.
\newline\urlprefix\url{http://dx.doi.org/10.1134/S1560354712030045}

\bibitem{Controlfuzzy2013}
E.~Kayacan, E.~Kayacan, H.~Ramon, W.~Saeys, Adaptive neuro-fuzzy control of a
  spherical rolling robot using sliding-mode-control-theory-based online
  learning algorithm, IEEE Trans. Cybern. 43~(1) (2013) 170--179.
\newblock \href {http://dx.doi.org/10.1109/TSMCB.2012.2202900}
  {\path{doi:10.1109/TSMCB.2012.2202900}}.

\bibitem{DORMAND1980ODE}
J.~Dormand, P.~Prince, A family of embedded runge-kutta formulae, J. Comput.
  Appl. Math. 6~(1) (1980) 19--26.
\newblock \href
  {http://dx.doi.org/http://dx.doi.org/10.1016/0771-050X(80)90013-3}
  {\path{doi:http://dx.doi.org/10.1016/0771-050X(80)90013-3}}.

\bibitem{Matlab1997ODE}
L.~F. Shampine, M.~W. Reichelt, The matlab ode suite, SIAM J. Scien. Comput.
  18~(1) (1997) 1--22.
\newblock \href {http://dx.doi.org/10.1137/S1064827594276424}
  {\path{doi:10.1137/S1064827594276424}}.

\bibitem{DoublePendulumMahboubi2012}
S.~Mahboubi, M.~Seyyed~Fakhrabadi, A.~Ghanbari, Design and implementation of a
  novel spherical mobile robot, J. Intell. Robot. Sys. 71~(1) (2012) 43--64.
\newblock \href {http://dx.doi.org/10.1007/s10846-012-9748-8}
  {\path{doi:10.1007/s10846-012-9748-8}}.

\end{thebibliography}


@article{RollingNature,
title = {Rolling in Nature and Robotics: A Review },
journal = {Journal of Bionic Engineering},
volume = {3},
number = {4},
pages = {195 - 208},
year = {2006},
note = {},
issn = {1672-6529},
doi = {10.1016/S1672-6529(07)60003-1},
author = {Rhodri H. Armour and Julian F.V. Vincent},
keywords = {Rolling robot},
}
@article{PolarNASA, 
author={A. Behar and J. Matthews and F. Carsey and J. Jones}, 
booktitle={Proc. IEEE Aerospace Conf.}, 
title={NASA/JPL Tumbleweed polar rover}, 
year={2004}, 
volume={1}, 
pages={395}, 
keywords={Earth;Mars;planetary rovers;planetary surfaces;Antarctic test;Antarctica;Earth;Greenland;Iridium satellite network;Jet Propulsion Laboratory;Mars;NASA;South Pole;Tumbleweed polar rover;global positioning system;ground penetrating radar;ground station;ice sheet;instrument payload;magnetometer;pressure data;remote settings;solar system bodies;temperature data;Antarctica;Earth;Environmental economics;Instruments;Laboratories;Mars;NASA;Payloads;Propulsion;Solar system}, 
doi={10.1109/AERO.2004.1367622}, 
ISSN={1095-323X}, 
month={March},}

@article{Tworigidrolling, 
author={Zexiang Li and Canny, John}, 
journal={IEEE Trans. Robot. Autom.}, 
title={Motion of two rigid bodies with rolling constraint}, 
year={1990}, 
volume={6}, 
number={1}, 
pages={62-72}, 
keywords={differential equations;matrix algebra;robots;Chow's theorem;Frobenius's theorem;admissible path;configuration space;contact configurations;contour following;differential equations;end-effector;matrix algebra;multifingered robot hand;rigid bodies;rolling constraint;wheeled mobile robotics;Cats;DC motors;Differential equations;Fingers;Friction;Geometry;Manipulators;Mobile robots;Motion control;Wheels}, 
doi={10.1109/70.88118}, 
ISSN={1042-296X}, 
month={Feb},}
@article{HalmeMotion1996, 
author={Halme, A. and Schonberg, T. and Yan Wang}, 
booktitle={Proc. 4th Int. Workshop Adv. Motion Control (AMC 1996)}, 
title={Motion control of a spherical mobile robot}, 
year={1996}, 
volume={1}, 
pages={259-264}, 
keywords={mobile robots;position control;robot dynamics;actuators;motion control;sensing devices;spherical mobile robot;stability;Actuators;Construction industry;Humans;Industrial control;Mobile robots;Motion control;Planets;Service robots;Stability;Turning}, 
doi={10.1109/AMC.1996.509415}, 
month={Mar},}
@article{BicchiSpherenonholonomy, 
author={Bicchi, A. and Balluchi, A. and Prattichizzo, D. and Gorelli, A.}, 
booktitle={Proc. IEEE Conf. Robot. Autom.}, 
title={Introducing the "SPHERICLE": an experimental testbed for research and teaching in nonholonomy}, 
year={1997}, 
volume={3}, 
pages={2620-2625}, 
keywords={mobile robots;nonlinear systems;path planning;robot dynamics;robot kinematics;student experiments;dynamics;experimental testbed;kinematics;motion planning;nonholonomic systems;plate-ball system;research;teaching;unicycle;untethered spherical vehicle;Control systems;Education;Kinematics;Laboratories;Mechanical systems;Mobile robots;Remotely operated vehicles;Testing;Underwater vehicles;Vehicle dynamics}, 
doi={10.1109/ROBOT.1997.619356}, 
month={Apr},}
@article{BrownCMUGyrosignlewheel, 
author={Brown, H.B., Jr. and Yangsheng Xu}, 
booktitle={in Proc. IEEE Int. Conf. on Robot. and Autom.}, 
title={A single-wheel, gyroscopically stabilized robot}, 
year={1996}, 
volume={4}, 
pages={3658-3663}, 
keywords={control system synthesis;gyroscopes;mechanical stability;mobile robots;robot dynamics;stability;Gyrover;dynamics;gyroscope;gyroscopically stabilized robot;manoeuvrability;mechanical stabilization;mobile robot;single-wheel vehicle;steering;Bicycles;Control systems;Gravity;Gyroscopes;Mobile robots;Motorcycles;Stability;Vehicle dynamics;Vehicles;Wheels}, 
doi={10.1109/ROBOT.1996.509270}, 
ISSN={1050-4729}, 
month={Apr},}
@article{Bhattacharyarolling2000, 
author={Bhattacharya, S. and Agrawal, S.K.}, 
journal={IEEE Trans. Robot. Autom.}, 
title={Spherical rolling robot: a design and motion planning studies}, 
year={2000}, 
volume={16}, 
number={6}, 
pages={835-839}, 
keywords={matrix algebra;mobile robots;path planning;telerobotics;time optimal control;feasible trajectories;minimum energy trajectories;minimum time trajectories;motion planning;nonholonomic constraints;nonholonomic robot system;overhead camera;remotely controlled internally mounted rotors;spherical rolling robot;Cameras;Mathematical model;Mechanical engineering;Mobile robots;Motion planning;Orbital robotics;Robot vision systems;Robotic assembly;Robotics and automation;Wheels}, 
doi={10.1109/70.897794}, 
ISSN={1042-296X}, 
month={Dec},}
@article{AugustJavadi2002, 
author={Amir Homayoun Javadi A and Mojabi, P.}, 
booktitle={in Proc. IEEE Int. Conf. on Robot. and Autom.}, 
title={Introducing August: a novel strategy for an omnidirectional spherical rolling robot}, 
year={2002}, 
volume={4}, 
pages={3527-3533}, 
keywords={acceleration control;mobile robots;path planning;position control;robot dynamics;velocity control;August;acceleration control;dynamics;internal propulsion;mobile robots;motion planning;omnidirectional robot;position control;spherical rolling robot;velocity control;Acceleration;Actuators;Mathematical model;Mobile robots;Motion planning;Propulsion;Skeleton;Stability;Trajectory;Wheels}, 
doi={10.1109/ROBOT.2002.1014256}, 
month={},}
@mastersthesis{MITSchroll2008,
  author        = {Schroll ,G. C.},
  title         = {Design of a Spherical Vehicle with Flywheel Momentum Storage for High Torque Capabilities},
  school        = {Massachusetts Institute of Technology},
  address       = {Cambridge},
  year          = {2008}
}
@article{DeformableElectro2011, 
author={Artusi, M. and Potz, M. and Aristizabal, J. and Menon, C. and Cocuzza, S. and Debei, S.}, 
journal={Mechatronics, IEEE/ASME Transactions on}, 
title={Electroactive Elastomeric Actuators for the Implementation of a Deformable Spherical Rover}, 
year={2011}, 
volume={16}, 
number={1}, 
pages={50-57}, 
keywords={deformation;elastomers;electroactive polymer actuators;electromechanical actuators;inflatable structures;manufacturing processes;planetary rovers;prototypes;dielectric elastomer actuator;electroactive elastomeric actuator;electromechanical behavior;inflated deformable rolling rover;locomotion system;prototype dynamics;spherical robot;Analytical models;Dielectrics;Force;Mathematical model;Prototypes;Shape;Stress;Concept design;dielectric elastomers;electroactive polymers;rover;smart materials;spherical robot}, 
doi={10.1109/TMECH.2010.2090163}, 
ISSN={1083-4435}, 
month={Feb},}

@article{HITdecoupled2006, 
author={Yue Ming and Deng Zongquan and Yu Xinyi and Yu Weizhen}, 
booktitle={Robotics and Biomimetics, 2006. ROBIO '06. IEEE International Conference on}, 
title={Introducing HIT Spherical Robot: Dynamic Modeling and Analysis Based on Decoupled Subsystem}, 
year={2006}, 
pages={181-186}, 
keywords={control engineering computing;control system analysis;mobile robots;motion control;open loop systems;robot dynamics;HIT spherical robot;decoupled subsystem;driving motion;dynamic complexity;dynamic equations;dynamic modeling;open-loop control system;real-time detection;Biomimetics;Control systems;Equations;Hardware;Mobile robots;Motion analysis;Motion detection;Open loop systems;Rotors;Turning;Spherical robot;decoupling principle;open-loop}, 
doi={10.1109/ROBIO.2006.340355}, 
month={Dec},}

@article{VolvotSMR2011, 
author={Ishikawa, M. and Kitayoshi, R. and Sugie, T.}, 
booktitle={IEEE Int. Conf. on Robot. and Biom. (ROBIO)}, 
title={Volvot : A spherical mobile robot with eccentric twin rotors}, 
year={2011}, 
pages={1462-1467}, 
keywords={actuators;drives;legged locomotion;numerical analysis;rotors;shafts;shells (structures);torque;2-DOF actuators;Volvot;angular acceleration;base actuator;drive shaft;driving force;eccentric rotors;eccentric twin rotors;gravity force;holonomy principle;mathematical dynamic model;numerical simulation;off-centered metal disks;periodic control scheme;reaction torque;rolling contact constraints;rotor actuator;spherical mobile robot;spherical outer shell;steady locomotion;uneven mass balance;Actuators;Gravity;Kinematics;Mobile robots;Rotors;Torque}, 
doi={10.1109/ROBIO.2011.6181496}, 
month={Dec},}
@article{Omnicron2012AIM, 
author={{Wei-Hsi} ,Chen and {Ching-Pei} ,Chen and {Wei-Shun} ,Yu and {Chang-Hao} ,Lin and {Pei-Chun} ,Lin}, 
booktitle={IEEE/ASME Int. Conf. Adv. Intell. Mechatron. (AIM)}, 
title={Design and implementation of an omnidirectional spherical robot Omnicron}, 
year={2012}, 
pages={719-724}, 
keywords={mobile robots;robot kinematics;trajectory control;wheels;3-degree-of-freedom planar omnidirectional mobility;forward 3-to-3 kinematic mapping;mobile robot;omnidirectional spherical robot Omnicron;omnidirectional wheels;spherical shell;trajectory-controllable mobility;Kinematics;Mobile robots;Robot kinematics;Robot sensing systems;Trajectory;Wheels}, 
doi={10.1109/AIM.2012.6266036}, 
ISSN={2159-6247}, 
month={July},}
@article{FirstPendulumBYQ2008, 
author={Daliang Liu and Hanxv Sun and Qingxuan Jia}, 
booktitle={Proc. IEEE Conf. Robot. Autom. and Mechatron.}, 
title={Stabilization and Path Following of a Spherical Robot}, 
year={2008}, 
pages={676-682}, 
keywords={asymptotic stability;mobile robots;motion control;nonlinear control systems;path planning;variable structure systems;driving torque;motion control;nonholonomic robot;path curvature controller;path following;planetary surface exploration;rolling robot;sliding-mode controller;spherical mobile robot;spherical robot stabilization;spinning angular velocity;three-dimensional nonlinear dynamic model;tracking errors;Angular velocity;Angular velocity control;Error correction;Mobile robots;Performance analysis;Security;Sliding mode control;Spinning;Stability analysis;Torque;dynamics;nonholonomic system;path following;sliding-mode control;spherical robot}, 
doi={10.1109/RAMECH.2008.4681358}, 
month={Sept},}


@article{DoublePendulumMahboubi2012,
author={Mahboubi, Saber
and Seyyed Fakhrabadi, {Mir Masoud}
and Ghanbari, Ahmad},
title={Design and Implementation of a Novel Spherical Mobile Robot},
journal={J. Intell. Robot. Sys.},
year={2012},
volume={71},
number={1},
pages={43--64},
abstract={In this paper, the design, modeling and implementation of a novel spherical mobile robot is presented. The robot composes of a spherical outer shell made of a transparent thermoplastic material, two pendulums, two DC motors with gearboxes, two equipments for linear motion and two control units. It possesses four distinct motional modes including: driving, steering, jumping and zero-radius turning. In driving and steering modes, the robot moves along straight and circular trajectories, respectively. The robot performs these motional modes using movable internal masses. In the jumping mode, it can jump over obstacles and in the zero-radius turning mode, the robot can turn with zero-radius to improve the motion flexibility. Furthermore, the attempts to establish the dynamic models of some motional modes are made and finally, the accuracy of the obtained dynamic models is verified by simulation and experimental results.},
issn={1573-0409},
doi={10.1007/s10846-012-9748-8}
}
@article{GeometryPhase2014, 
author={A. Morinaga and M. Svinin and M. Yamamoto}, 
journal={IEEE Trans. Robot.}, 
title={A Motion Planning Strategy for a Spherical Rolling Robot Driven by Two Internal Rotors}, 
year={2014}, 
volume={30}, 
number={4}, 
pages={993-1002}, 
keywords={differential geometry;iterative methods;path planning;robot dynamics;rotors (mechanical);contact trajectory;dynamic formulation;equatorial line;geodesic line;geometric phase;internal rotors;iterative steering algorithm;motion planning strategy;nilpotent approximation;nonnilpotent robot dynamics;nontrivial maneuver;orthogonal axes;singularity line;spherical rolling robot;trivial maneuver;Approximation methods;Dynamics;Heuristic algorithms;Planning;Robots;Rotors;Vectors;Dynamics;motion planning;nonholonomic systems;rolling constraints;spherical rolling robot}, 
doi={10.1109/TRO.2014.2307112}, 
ISSN={1552-3098}, 
month={Aug},}


@article{Algorithmbase2010, 
author={F. Alouges and Y. Chitour and R. Long}, 
journal={IEEE Trans. Robot.}, 
title={A Motion-Planning Algorithm for the Rolling-Body Problem}, 
year={2010}, 
volume={26}, 
number={5}, 
pages={827-836}, 
keywords={manipulators;path planning;constructive planning algorithm;continuation method;convergence speed;motion planning algorithm;numerical implementation;rolling body problem;strictly convex surface rolling;Aerospace electronics;Algorithm design and analysis;Control systems;Controllability;Convergence;Equations;Manifolds;Manipulators;Motion planning;Motion-planning;Planning;Robots;Robustness;Spinning;Trajectory;Manipulation planning;nonholonomic motion planning}, 
doi={10.1109/TRO.2010.2053733}, 
ISSN={1552-3098}, 
month={Oct},}


@article{Controlfuzzy2013, 
author={E. Kayacan and E. Kayacan and H. Ramon and W. Saeys}, 
journal={IEEE Trans. Cybern.}, 
title={Adaptive Neuro-Fuzzy Control of a Spherical Rolling Robot Using Sliding-Mode-Control-Theory-Based Online Learning Algorithm}, 
year={2013}, 
volume={43}, 
number={1}, 
pages={170-179}, 
keywords={Lyapunov methods;adaptive control;asymptotic stability;fuzzy control;fuzzy neural nets;learning systems;mobile robots;neurocontrollers;robot dynamics;transient response;variable structure systems;Lyapunov function;SMC theory;SMC-theory-based learning algorithm;adaptive neuro-fuzzy controller;asymptotic stability;control structure;conventional controller;dynamic equations;learning stability;neuro-fuzzy network;neuro-fuzzy system;parameter updating rules;parameter variations;real system;sliding-mode-control-theory-based online learning algorithm;spherical rolling robot;steady-state error;transient response performance;unmodeled dynamics;Equations;Fuzzy control;Fuzzy neural networks;Heuristic algorithms;Mathematical model;Mobile robots;Adaptive neuro-fuzzy control;sliding-mode learning algorithm;spherical rolling robot}, 
doi={10.1109/TSMCB.2012.2202900}, 
ISSN={2168-2267}, 
month={Feb},}

@article{GeometryPhase2005, 
author={G. Oriolo and M. Vendittelli}, 
journal={IEEE Trans. Robot.}, 
title={A framework for the stabilization of general nonholonomic systems with an application to the plate-ball mechanism}, 
year={2005}, 
volume={21}, 
number={2}, 
pages={162-175}, 
keywords={asymptotic stability;manipulator dynamics;approximate steering control;asymptotic stability;exponential convergence;general nonholonomic systems stabilization;iterative control scheme;nilpotent system dynamics approximation;plate-ball mechanism;Asymptotic stability;Control systems;Convergence;Feedback;Iterative methods;Kinematics;Mechanical factors;Mobile robots;Orbital robotics;Robust control;Iterative steering (IS);nilpotent approximations (NAs);nonholonomic systems;plate-ball mechanism;stabilization}, 
doi={10.1109/TRO.2004.839231}, 
ISSN={1552-3098}, 
month={April},}

 
@article{Algorithmbase2008, 
author={M. Svinin and S. Hosoe}, 
journal={IEEE Trans. Robot.}, 
title={Motion Planning Algorithms for a Rolling Sphere With Limited Contact Area}, 
year={2008}, 
volume={24}, 
number={3}, 
pages={612-625}, 
keywords={manipulators;path planning;generalized Viviani-curve-based ones;hemispherical object;limited contact area;motion planning algorithms;parallel parking;rolling sphere;Motion planning;nonholonomic systems;optimality;rolling constraints}, 
doi={10.1109/TRO.2008.921568}, 
ISSN={1552-3098}, 
month={June},}

@article{GeometryPhase2015, 
author={V. Muralidharan and A. D. Mahindrakar}, 
journal={IEEE Trans. Automat. Control}, 
title={Geometric Controllability and Stabilization of Spherical Robot Dynamics}, 
year={2015}, 
volume={60}, 
number={10}, 
pages={2762-2767}, 
keywords={asymptotic stability;controllability;mobile robots;nonlinear control systems;robot dynamics;Brockett condition;asymptotically stabilizing geometric control law;geometric controllability;independent inertia disc actuators;nonholonomic spherical robot;position trajectory tracking;smooth global tracking controller;spherical robot dynamics;stabilization;Asymptotic stability;Attitude control;Controllability;Kinematics;Robots;Trajectory;Vectors;Algebraic/geometric methods;position and attitude stabilization;robotics;spherical robot;stability of NL systems}, 
doi={10.1109/TAC.2015.2404512}, 
ISSN={0018-9286}, 
month={Oct},}


@article{Controlnonholonomic2012,
author="Borisov, Alexey V.
and Kilin, Alexander A.
and Mamaev, Ivan S.",
title="How to control Chaplygin's sphere using rotors",
journal="Reg. and Chaotic Dynam.",
year="2012",
volume="17",
number="3",
pages="258--272",
abstract="In the paper we study the control of a balanced dynamically non-symmetric sphere with rotors. The no-slip condition at the point of contact is assumed. The algebraic controllability is shown and the control inputs that steer the ball along a given trajectory on the plane are found. For some simple trajectories explicit tracking algorithms are proposed.",
issn="1468-4845",
doi="10.1134/S1560354712030045",
url="http://dx.doi.org/10.1134/S1560354712030045"
}
@article{DORMAND1980ODE,
title = {A family of embedded Runge-Kutta formulae},
journal = {J. Comput. Appl. Math.},
volume = {6},
number = {1},
pages = {19-26},
year = {1980},
issn = {0377-0427},
doi = {http://dx.doi.org/10.1016/0771-050X(80)90013-3},
author = {J.R. Dormand and P.J. Prince}
}
@article{Matlab1997ODE,
author = {Lawrence F. Shampine and Mark W. Reichelt},
title = {The MATLAB ODE Suite},
journal = {SIAM J. Scien. Comput.},
volume = {18},
number = {1},
pages = {1-22},
year = {1997},
doi = {10.1137/S1064827594276424}
}

@article{Moball2015,
author={J. Asama and M. R. Burkhardt and F. Davoodi and J. W. Burdick},
booktitle={in Proc. IEEE Int. Conf. on Robot. and Autom.},
title={Design investigation of a coreless tubular linear generator for a Moball: A spherical exploration robot with wind-energy harvesting capability},
year={2015},
pages={244-251},
keywords={energy harvesting;finite element analysis;linear machines;mobile robots;power engineering computing;wind power;Moball;coreless tubular linear electromagnetic generator;dipole PM;energy generation improvement;energy scavenging system;finite element analysis;load resistance;motion control;slope-shaped back-iron;wind energy harvesting capability;wind-driven spherical exploration robot;Energy harvesting;Generators;Iron;Magnetic flux;Robots;Solenoids;Stator cores},
doi={10.1109/ICRA.2015.7139007},
ISSN={1050-4729},
month={May},}

@article{InclinedPYU2013,
author={{T. Yu} and {H. Sun} and {Q. Jia} and {Y. Zhang} and {W. Zhao}},
booktitle={Res. J. Appl. Sci. Eng. Technology},
title={Stabilization and Control of a Spherical Robot
on an Inclined Plane},
year={2013},
pages={2289–-2296},
ISSN={2289–2296},
keywords={inclined plain},
month={May},}

@article{Chen201335,
title = {Design and implementation of a ball-driven omnidirectional spherical robot },
journal = {Mechanism and Machine Theory },
volume = {68},
year = {2013},
pages = {35 - 48},
issn = {0094-114X},
doi = {http://dx.doi.org/10.1016/j.mechmachtheory.2013.04.012},
author = {Wei-Hsi Chen and Ching-Pei Chen and Jia-Shiuan Tsai and Jackie Yang and Pei-Chun Lin},
keywords = {Spherical robot},
keywords = {Omnidirectional},
keywords = {Locomotion}
}

@article{Hogan2015,
author={Hogan, Fran{\c{c}}ois Robert
and Forbes, James Richard},
title={Modeling of spherical robots rolling on generic surfaces},
journal={Multibody System Dynamics},
year={2015},
volume={35},
number={1},
pages={91--109},
issn={1573-272X},
doi={10.1007/s11044-014-9438-3},
}



author={{S. A. Tafrishi}},
title={{"RollRoller"} Novel Spherical Mobile Robot Basic Dynamical Analysis and Motion Simulations},
school        = {University of Sheffield},
address       = {Sheffield, UK},
year          = {2014},
month = {Sep.}
}







@article{RollingNature,
title = {Rolling in Nature and Robotics: A Review },
journal = {Journal of Bionic Engineering},
volume = {3},
number = {4},
pages = {195 - 208},
year = {2006},
note = {},
issn = {1672-6529},
doi = {10.1016/S1672-6529(07)60003-1},
author = {Rhodri H. Armour and Julian F.V. Vincent},
keywords = {Rolling robot},
}
@article{PolarNASA, 
author={A. Behar and J. Matthews and F. Carsey and J. Jones}, 
booktitle={Proc. IEEE Aerospace Conf.}, 
title={NASA/JPL Tumbleweed polar rover}, 
year={2004}, 
volume={1}, 
pages={395}, 
keywords={Earth;Mars;planetary rovers;planetary surfaces;Antarctic test;Antarctica;Earth;Greenland;Iridium satellite network;Jet Propulsion Laboratory;Mars;NASA;South Pole;Tumbleweed polar rover;global positioning system;ground penetrating radar;ground station;ice sheet;instrument payload;magnetometer;pressure data;remote settings;solar system bodies;temperature data;Antarctica;Earth;Environmental economics;Instruments;Laboratories;Mars;NASA;Payloads;Propulsion;Solar system}, 
doi={10.1109/AERO.2004.1367622}, 
ISSN={1095-323X}, 
month={March},}

@article{Tworigidrolling, 
author={Zexiang Li and Canny, John}, 
journal={IEEE Trans. Robot. Autom.}, 
title={Motion of two rigid bodies with rolling constraint}, 
year={1990}, 
volume={6}, 
number={1}, 
pages={62-72}, 
keywords={differential equations;matrix algebra;robots;Chow's theorem;Frobenius's theorem;admissible path;configuration space;contact configurations;contour following;differential equations;end-effector;matrix algebra;multifingered robot hand;rigid bodies;rolling constraint;wheeled mobile robotics;Cats;DC motors;Differential equations;Fingers;Friction;Geometry;Manipulators;Mobile robots;Motion control;Wheels}, 
doi={10.1109/70.88118}, 
ISSN={1042-296X}, 
month={Feb},}
@article{HalmeMotion1996, 
author={Halme, A. and Schonberg, T. and Yan Wang}, 
booktitle={Proc. 4th Int. Workshop Adv. Motion Control (AMC 1996)}, 
title={Motion control of a spherical mobile robot}, 
year={1996}, 
volume={1}, 
pages={259-264}, 
keywords={mobile robots;position control;robot dynamics;actuators;motion control;sensing devices;spherical mobile robot;stability;Actuators;Construction industry;Humans;Industrial control;Mobile robots;Motion control;Planets;Service robots;Stability;Turning}, 
doi={10.1109/AMC.1996.509415}, 
month={Mar},}
@article{BicchiSpherenonholonomy, 
author={Bicchi, A. and Balluchi, A. and Prattichizzo, D. and Gorelli, A.}, 
booktitle={Proc. IEEE Conf. Robot. Autom.}, 
title={Introducing the "SPHERICLE": an experimental testbed for research and teaching in nonholonomy}, 
year={1997}, 
volume={3}, 
pages={2620-2625}, 
keywords={mobile robots;nonlinear systems;path planning;robot dynamics;robot kinematics;student experiments;dynamics;experimental testbed;kinematics;motion planning;nonholonomic systems;plate-ball system;research;teaching;unicycle;untethered spherical vehicle;Control systems;Education;Kinematics;Laboratories;Mechanical systems;Mobile robots;Remotely operated vehicles;Testing;Underwater vehicles;Vehicle dynamics}, 
doi={10.1109/ROBOT.1997.619356}, 
month={Apr},}
@article{BrownCMUGyrosignlewheel, 
author={Brown, H.B., Jr. and Yangsheng Xu}, 
booktitle={in Proc. IEEE Int. Conf. on Robot. and Autom.}, 
title={A single-wheel, gyroscopically stabilized robot}, 
year={1996}, 
volume={4}, 
pages={3658-3663}, 
keywords={control system synthesis;gyroscopes;mechanical stability;mobile robots;robot dynamics;stability;Gyrover;dynamics;gyroscope;gyroscopically stabilized robot;manoeuvrability;mechanical stabilization;mobile robot;single-wheel vehicle;steering;Bicycles;Control systems;Gravity;Gyroscopes;Mobile robots;Motorcycles;Stability;Vehicle dynamics;Vehicles;Wheels}, 
doi={10.1109/ROBOT.1996.509270}, 
ISSN={1050-4729}, 
month={Apr},}
@article{Bhattacharyarolling2000, 
author={Bhattacharya, S. and Agrawal, S.K.}, 
journal={IEEE Trans. Robot. Autom.}, 
title={Spherical rolling robot: a design and motion planning studies}, 
year={2000}, 
volume={16}, 
number={6}, 
pages={835-839}, 
keywords={matrix algebra;mobile robots;path planning;telerobotics;time optimal control;feasible trajectories;minimum energy trajectories;minimum time trajectories;motion planning;nonholonomic constraints;nonholonomic robot system;overhead camera;remotely controlled internally mounted rotors;spherical rolling robot;Cameras;Mathematical model;Mechanical engineering;Mobile robots;Motion planning;Orbital robotics;Robot vision systems;Robotic assembly;Robotics and automation;Wheels}, 
doi={10.1109/70.897794}, 
ISSN={1042-296X}, 
month={Dec},}
@article{AugustJavadi2002, 
author={Amir Homayoun Javadi A and Mojabi, P.}, 
booktitle={in Proc. IEEE Int. Conf. on Robot. and Autom.}, 
title={Introducing August: a novel strategy for an omnidirectional spherical rolling robot}, 
year={2002}, 
volume={4}, 
pages={3527-3533}, 
keywords={acceleration control;mobile robots;path planning;position control;robot dynamics;velocity control;August;acceleration control;dynamics;internal propulsion;mobile robots;motion planning;omnidirectional robot;position control;spherical rolling robot;velocity control;Acceleration;Actuators;Mathematical model;Mobile robots;Motion planning;Propulsion;Skeleton;Stability;Trajectory;Wheels}, 
doi={10.1109/ROBOT.2002.1014256}, 
month={},}
@mastersthesis{MITSchroll2008,
  author        = {Schroll ,G. C.},
  title         = {Design of a Spherical Vehicle with Flywheel Momentum Storage for High Torque Capabilities},
  school        = {Massachusetts Institute of Technology},
  address       = {Cambridge},
  year          = {2008}
}
@article{DeformableElectro2011, 
author={Artusi, M. and Potz, M. and Aristizabal, J. and Menon, C. and Cocuzza, S. and Debei, S.}, 
journal={Mechatronics, IEEE/ASME Transactions on}, 
title={Electroactive Elastomeric Actuators for the Implementation of a Deformable Spherical Rover}, 
year={2011}, 
volume={16}, 
number={1}, 
pages={50-57}, 
keywords={deformation;elastomers;electroactive polymer actuators;electromechanical actuators;inflatable structures;manufacturing processes;planetary rovers;prototypes;dielectric elastomer actuator;electroactive elastomeric actuator;electromechanical behavior;inflated deformable rolling rover;locomotion system;prototype dynamics;spherical robot;Analytical models;Dielectrics;Force;Mathematical model;Prototypes;Shape;Stress;Concept design;dielectric elastomers;electroactive polymers;rover;smart materials;spherical robot}, 
doi={10.1109/TMECH.2010.2090163}, 
ISSN={1083-4435}, 
month={Feb},}

@article{HITdecoupled2006, 
author={Yue Ming and Deng Zongquan and Yu Xinyi and Yu Weizhen}, 
booktitle={Robotics and Biomimetics, 2006. ROBIO '06. IEEE International Conference on}, 
title={Introducing HIT Spherical Robot: Dynamic Modeling and Analysis Based on Decoupled Subsystem}, 
year={2006}, 
pages={181-186}, 
keywords={control engineering computing;control system analysis;mobile robots;motion control;open loop systems;robot dynamics;HIT spherical robot;decoupled subsystem;driving motion;dynamic complexity;dynamic equations;dynamic modeling;open-loop control system;real-time detection;Biomimetics;Control systems;Equations;Hardware;Mobile robots;Motion analysis;Motion detection;Open loop systems;Rotors;Turning;Spherical robot;decoupling principle;open-loop}, 
doi={10.1109/ROBIO.2006.340355}, 
month={Dec},}

@article{VolvotSMR2011, 
author={Ishikawa, M. and Kitayoshi, R. and Sugie, T.}, 
booktitle={IEEE Int. Conf. on Robot. and Biom. (ROBIO)}, 
title={Volvot : A spherical mobile robot with eccentric twin rotors}, 
year={2011}, 
pages={1462-1467}, 
keywords={actuators;drives;legged locomotion;numerical analysis;rotors;shafts;shells (structures);torque;2-DOF actuators;Volvot;angular acceleration;base actuator;drive shaft;driving force;eccentric rotors;eccentric twin rotors;gravity force;holonomy principle;mathematical dynamic model;numerical simulation;off-centered metal disks;periodic control scheme;reaction torque;rolling contact constraints;rotor actuator;spherical mobile robot;spherical outer shell;steady locomotion;uneven mass balance;Actuators;Gravity;Kinematics;Mobile robots;Rotors;Torque}, 
doi={10.1109/ROBIO.2011.6181496}, 
month={Dec},}
@article{Omnicron2012AIM, 
author={{Wei-Hsi} ,Chen and {Ching-Pei} ,Chen and {Wei-Shun} ,Yu and {Chang-Hao} ,Lin and {Pei-Chun} ,Lin}, 
booktitle={IEEE/ASME Int. Conf. Adv. Intell. Mechatron. (AIM)}, 
title={Design and implementation of an omnidirectional spherical robot Omnicron}, 
year={2012}, 
pages={719-724}, 
keywords={mobile robots;robot kinematics;trajectory control;wheels;3-degree-of-freedom planar omnidirectional mobility;forward 3-to-3 kinematic mapping;mobile robot;omnidirectional spherical robot Omnicron;omnidirectional wheels;spherical shell;trajectory-controllable mobility;Kinematics;Mobile robots;Robot kinematics;Robot sensing systems;Trajectory;Wheels}, 
doi={10.1109/AIM.2012.6266036}, 
ISSN={2159-6247}, 
month={July},}
@article{FirstPendulumBYQ2008, 
author={Daliang Liu and Hanxv Sun and Qingxuan Jia}, 
booktitle={Proc. IEEE Conf. Robot. Autom. and Mechatron.}, 
title={Stabilization and Path Following of a Spherical Robot}, 
year={2008}, 
pages={676-682}, 
keywords={asymptotic stability;mobile robots;motion control;nonlinear control systems;path planning;variable structure systems;driving torque;motion control;nonholonomic robot;path curvature controller;path following;planetary surface exploration;rolling robot;sliding-mode controller;spherical mobile robot;spherical robot stabilization;spinning angular velocity;three-dimensional nonlinear dynamic model;tracking errors;Angular velocity;Angular velocity control;Error correction;Mobile robots;Performance analysis;Security;Sliding mode control;Spinning;Stability analysis;Torque;dynamics;nonholonomic system;path following;sliding-mode control;spherical robot}, 
doi={10.1109/RAMECH.2008.4681358}, 
month={Sept},}


@article{DoublePendulumMahboubi2012,
author={Mahboubi, Saber
and Seyyed Fakhrabadi, {Mir Masoud}
and Ghanbari, Ahmad},
title={Design and Implementation of a Novel Spherical Mobile Robot},
journal={J. Intell. Robot. Sys.},
year={2012},
volume={71},
number={1},
pages={43--64},
abstract={In this paper, the design, modeling and implementation of a novel spherical mobile robot is presented. The robot composes of a spherical outer shell made of a transparent thermoplastic material, two pendulums, two DC motors with gearboxes, two equipments for linear motion and two control units. It possesses four distinct motional modes including: driving, steering, jumping and zero-radius turning. In driving and steering modes, the robot moves along straight and circular trajectories, respectively. The robot performs these motional modes using movable internal masses. In the jumping mode, it can jump over obstacles and in the zero-radius turning mode, the robot can turn with zero-radius to improve the motion flexibility. Furthermore, the attempts to establish the dynamic models of some motional modes are made and finally, the accuracy of the obtained dynamic models is verified by simulation and experimental results.},
issn={1573-0409},
doi={10.1007/s10846-012-9748-8}
}
@article{GeometryPhase2014, 
author={A. Morinaga and M. Svinin and M. Yamamoto}, 
journal={IEEE Trans. Robot.}, 
title={A Motion Planning Strategy for a Spherical Rolling Robot Driven by Two Internal Rotors}, 
year={2014}, 
volume={30}, 
number={4}, 
pages={993-1002}, 
keywords={differential geometry;iterative methods;path planning;robot dynamics;rotors (mechanical);contact trajectory;dynamic formulation;equatorial line;geodesic line;geometric phase;internal rotors;iterative steering algorithm;motion planning strategy;nilpotent approximation;nonnilpotent robot dynamics;nontrivial maneuver;orthogonal axes;singularity line;spherical rolling robot;trivial maneuver;Approximation methods;Dynamics;Heuristic algorithms;Planning;Robots;Rotors;Vectors;Dynamics;motion planning;nonholonomic systems;rolling constraints;spherical rolling robot}, 
doi={10.1109/TRO.2014.2307112}, 
ISSN={1552-3098}, 
month={Aug},}


@article{Algorithmbase2010, 
author={F. Alouges and Y. Chitour and R. Long}, 
journal={IEEE Trans. Robot.}, 
title={A Motion-Planning Algorithm for the Rolling-Body Problem}, 
year={2010}, 
volume={26}, 
number={5}, 
pages={827-836}, 
keywords={manipulators;path planning;constructive planning algorithm;continuation method;convergence speed;motion planning algorithm;numerical implementation;rolling body problem;strictly convex surface rolling;Aerospace electronics;Algorithm design and analysis;Control systems;Controllability;Convergence;Equations;Manifolds;Manipulators;Motion planning;Motion-planning;Planning;Robots;Robustness;Spinning;Trajectory;Manipulation planning;nonholonomic motion planning}, 
doi={10.1109/TRO.2010.2053733}, 
ISSN={1552-3098}, 
month={Oct},}


@article{Controlfuzzy2013, 
author={E. Kayacan and E. Kayacan and H. Ramon and W. Saeys}, 
journal={IEEE Trans. Cybern.}, 
title={Adaptive Neuro-Fuzzy Control of a Spherical Rolling Robot Using Sliding-Mode-Control-Theory-Based Online Learning Algorithm}, 
year={2013}, 
volume={43}, 
number={1}, 
pages={170-179}, 
keywords={Lyapunov methods;adaptive control;asymptotic stability;fuzzy control;fuzzy neural nets;learning systems;mobile robots;neurocontrollers;robot dynamics;transient response;variable structure systems;Lyapunov function;SMC theory;SMC-theory-based learning algorithm;adaptive neuro-fuzzy controller;asymptotic stability;control structure;conventional controller;dynamic equations;learning stability;neuro-fuzzy network;neuro-fuzzy system;parameter updating rules;parameter variations;real system;sliding-mode-control-theory-based online learning algorithm;spherical rolling robot;steady-state error;transient response performance;unmodeled dynamics;Equations;Fuzzy control;Fuzzy neural networks;Heuristic algorithms;Mathematical model;Mobile robots;Adaptive neuro-fuzzy control;sliding-mode learning algorithm;spherical rolling robot}, 
doi={10.1109/TSMCB.2012.2202900}, 
ISSN={2168-2267}, 
month={Feb},}

@article{GeometryPhase2005, 
author={G. Oriolo and M. Vendittelli}, 
journal={IEEE Trans. Robot.}, 
title={A framework for the stabilization of general nonholonomic systems with an application to the plate-ball mechanism}, 
year={2005}, 
volume={21}, 
number={2}, 
pages={162-175}, 
keywords={asymptotic stability;manipulator dynamics;approximate steering control;asymptotic stability;exponential convergence;general nonholonomic systems stabilization;iterative control scheme;nilpotent system dynamics approximation;plate-ball mechanism;Asymptotic stability;Control systems;Convergence;Feedback;Iterative methods;Kinematics;Mechanical factors;Mobile robots;Orbital robotics;Robust control;Iterative steering (IS);nilpotent approximations (NAs);nonholonomic systems;plate-ball mechanism;stabilization}, 
doi={10.1109/TRO.2004.839231}, 
ISSN={1552-3098}, 
month={April},}

 
@article{Algorithmbase2008, 
author={M. Svinin and S. Hosoe}, 
journal={IEEE Trans. Robot.}, 
title={Motion Planning Algorithms for a Rolling Sphere With Limited Contact Area}, 
year={2008}, 
volume={24}, 
number={3}, 
pages={612-625}, 
keywords={manipulators;path planning;generalized Viviani-curve-based ones;hemispherical object;limited contact area;motion planning algorithms;parallel parking;rolling sphere;Motion planning;nonholonomic systems;optimality;rolling constraints}, 
doi={10.1109/TRO.2008.921568}, 
ISSN={1552-3098}, 
month={June},}

@article{GeometryPhase2015, 
author={V. Muralidharan and A. D. Mahindrakar}, 
journal={IEEE Trans. Automat. Control}, 
title={Geometric Controllability and Stabilization of Spherical Robot Dynamics}, 
year={2015}, 
volume={60}, 
number={10}, 
pages={2762-2767}, 
keywords={asymptotic stability;controllability;mobile robots;nonlinear control systems;robot dynamics;Brockett condition;asymptotically stabilizing geometric control law;geometric controllability;independent inertia disc actuators;nonholonomic spherical robot;position trajectory tracking;smooth global tracking controller;spherical robot dynamics;stabilization;Asymptotic stability;Attitude control;Controllability;Kinematics;Robots;Trajectory;Vectors;Algebraic/geometric methods;position and attitude stabilization;robotics;spherical robot;stability of NL systems}, 
doi={10.1109/TAC.2015.2404512}, 
ISSN={0018-9286}, 
month={Oct},}


@article{Controlnonholonomic2012,
author="Borisov, Alexey V.
and Kilin, Alexander A.
and Mamaev, Ivan S.",
title="How to control Chaplygin's sphere using rotors",
journal="Reg. and Chaotic Dynam.",
year="2012",
volume="17",
number="3",
pages="258--272",
abstract="In the paper we study the control of a balanced dynamically non-symmetric sphere with rotors. The no-slip condition at the point of contact is assumed. The algebraic controllability is shown and the control inputs that steer the ball along a given trajectory on the plane are found. For some simple trajectories explicit tracking algorithms are proposed.",
issn="1468-4845",
doi="10.1134/S1560354712030045",
url="http://dx.doi.org/10.1134/S1560354712030045"
}
@article{DORMAND1980ODE,
title = {A family of embedded Runge-Kutta formulae},
journal = {J. Comput. Appl. Math.},
volume = {6},
number = {1},
pages = {19-26},
year = {1980},
issn = {0377-0427},
doi = {http://dx.doi.org/10.1016/0771-050X(80)90013-3},
author = {J.R. Dormand and P.J. Prince}
}
@article{Matlab1997ODE,
author = {Lawrence F. Shampine and Mark W. Reichelt},
title = {The MATLAB ODE Suite},
journal = {SIAM J. Scien. Comput.},
volume = {18},
number = {1},
pages = {1-22},
year = {1997},
doi = {10.1137/S1064827594276424}
}

@article{Moball2015,
author={J. Asama and M. R. Burkhardt and F. Davoodi and J. W. Burdick},
booktitle={in Proc. IEEE Int. Conf. on Robot. and Autom.},
title={Design investigation of a coreless tubular linear generator for a Moball: A spherical exploration robot with wind-energy harvesting capability},
year={2015},
pages={244-251},
keywords={energy harvesting;finite element analysis;linear machines;mobile robots;power engineering computing;wind power;Moball;coreless tubular linear electromagnetic generator;dipole PM;energy generation improvement;energy scavenging system;finite element analysis;load resistance;motion control;slope-shaped back-iron;wind energy harvesting capability;wind-driven spherical exploration robot;Energy harvesting;Generators;Iron;Magnetic flux;Robots;Solenoids;Stator cores},
doi={10.1109/ICRA.2015.7139007},
ISSN={1050-4729},
month={May},}

@article{InclinedPYU2013,
author={{T. Yu} and {H. Sun} and {Q. Jia} and {Y. Zhang} and {W. Zhao}},
booktitle={Res. J. Appl. Sci. Eng. Technology},
title={Stabilization and Control of a Spherical Robot
on an Inclined Plane},
year={2013},
pages={2289–-2296},
ISSN={2289–2296},
keywords={inclined plain},
month={May},}

@article{Chen201335,
title = {Design and implementation of a ball-driven omnidirectional spherical robot },
journal = {Mechanism and Machine Theory },
volume = {68},
year = {2013},
pages = {35 - 48},
issn = {0094-114X},
doi = {http://dx.doi.org/10.1016/j.mechmachtheory.2013.04.012},
author = {Wei-Hsi Chen and Ching-Pei Chen and Jia-Shiuan Tsai and Jackie Yang and Pei-Chun Lin},
keywords = {Spherical robot},
keywords = {Omnidirectional},
keywords = {Locomotion}
}

@article{Hogan2015,
author={Hogan, Fran{\c{c}}ois Robert
and Forbes, James Richard},
title={Modeling of spherical robots rolling on generic surfaces},
journal={Multibody System Dynamics},
year={2015},
volume={35},
number={1},
pages={91--109},
issn={1573-272X},
doi={10.1007/s11044-014-9438-3},
}



author={{S. A. Tafrishi}},
title={{"RollRoller"} Novel Spherical Mobile Robot Basic Dynamical Analysis and Motion Simulations},
school        = {University of Sheffield},
address       = {Sheffield, UK},
year          = {2014},
month = {Sep.}
}

\end{document}